\documentclass[final,3p,times]{elsarticle}

\usepackage{hyperref}
\usepackage{subfigure}
\usepackage{calc}
\usepackage{amsmath}
\usepackage{amstext}
\usepackage{amssymb}
\usepackage{multicol}
\usepackage{multirow}
\usepackage{color}
\usepackage{colortbl}
\usepackage{hhline}

\definecolor{LightGray}{gray}{0.8}
\newcolumntype{g}{>{\columncolor{LightGray}}c}
\newcommand{\gc}[1]{\multicolumn{1}{g}{#1}} % grey inner cell
\newcommand{\gcr}[1]{\multicolumn{1}{g|}{#1}} % grey right cell

\newcommand\toppad{\rule{0pt}{2.4ex}}

\def\argmax{\mathop{\mathrm{argmax}\;}\limits}

\journal{Applied Soft Computing}

\begin{document}

\begin{frontmatter}

\title{Constructing Parsimonious Analytic Models for Dynamic Systems via Symbolic Regression}

\author[a1,a2]{Erik Derner}
\author[a1]{Ji\v{r}\'i Kubal\'ik}
\author[a3]{Nicola Ancona}
\author[a1,a3]{Robert Babu\v{s}ka}

\address[a1]{Czech Institute of Informatics, Robotics, and Cybernetics, Czech Technical University in Prague, Prague, 16000, Czech Republic}
\address[a2]{Department of Control Engineering, Faculty of Electrical Engineering, Czech Technical University in Prague, Prague, 12135, Czech Republic}
\address[a3]{Cognitive Robotics, Delft University of Technology, Delft, 2628 CD, The Netherlands}
\address{Corresponding author: Erik Derner, \texttt{erik.derner@cvut.cz}}

\begin{abstract}
Developing mathematical models of dynamic systems is central to many disciplines of engineering and science. Models facilitate simulations, analysis of the system's behavior, decision making and design of automatic control algorithms. Even inherently model-free control techniques such as reinforcement learning (RL) have been shown to benefit from the use of models, typically learned online. Any model construction method must address the tradeoff between the accuracy of the model and its complexity, which is difficult to strike. In this paper, we propose to employ symbolic regression (SR) to construct parsimonious process models described by analytic equations. We have equipped our method with two different state-of-the-art SR algorithms which automatically search for equations that fit the measured data: Single Node Genetic Programming (SNGP) and Multi-Gene Genetic Programming (MGGP). In addition to the standard problem formulation in the state-space domain, we show how the method can also be applied to input--output models of the NARX (nonlinear autoregressive with exogenous input) type. We present the approach on three simulated examples with up to 14-dimensional state space: an inverted pendulum, a mobile robot, and a bipedal walking robot. A comparison with deep neural networks and local linear regression shows that SR in most cases outperforms these commonly used alternative methods. We demonstrate on a real pendulum system that the analytic model found enables a RL controller to successfully perform the swing-up task, based on a model constructed from only 100 data samples.
\end{abstract}

%\begin{highlights}
%\item Accurate models can be found from small data sets and used for real-time control
%\item Symbolic regression finds both the structure and the parameters of the models
%\item The models have the form of mathematical expressions, facilitating further processing
%\item The method works both with state-space and input--output (NARX-type) models
%\item Extensive experimental evaluation on three systems demonstrates its practical utility
%\end{highlights}


\end{frontmatter}

%% main text
\section{Introduction}
\label{sec:introduction}

Numerous methods rely on an accurate model of the system. Model-based techniques comprise a wide variety of methods such as model predictive control \cite{rawlings2009model,mayne2000constrained}, time series prediction \cite{masters1995neural}, fault detection and diagnosis \cite{gertler2013fault,venkatasubramanian2003review}, or reinforcement learning (RL) \cite{sutton2018reinforcement,Mnih15}.

Even though model-free algorithms are available, the absence of a model slows down convergence and leads to extensive learning times \cite{Shixiang16,Peters06,Kober12}.
Various model-based methods have been proposed to speed up learning \cite{Kuvayev96,Forbes02,Hester17,Jong07,Sutton91}. To that end, many model-learning approaches are available: time-varying linear models \cite{Levine14,Lioutikov14}, Gaussian processes \cite{Deisenroth11,Boedecker14} and other probabilistic models \cite{Ng06}, basis function expansions \cite{munos2002variable,Busoniu11-SMC}, regression trees \cite{ernst2005tree}, deep neural networks \cite{Lange12,Mnih13,Mnih15,Lillicrap15,Heess15,deBruin18-RAL,deBruin18-JMLR} or local linear regression \cite{Atkeson1997,Grondman12-SMC,Lieck16}.

All the above approaches suffer from drawbacks induced by the use of the specific approximation technique, such as a large number of parameters (deep neural networks), local nature of the approximator (local linear regression), computational complexity (Gaussian processes), etc. In this article, we propose another way to capture the system dynamics: using analytic models constructed by means of the symbolic regression method (SR). Symbolic regression is based on genetic programming and it has been used in nonlinear data-driven modeling, often with quite impressive results \cite{schmidt2009distilling,staelens2012constructing,brauer2012using,vladislavleva2013predicting,Zegklitz17}.

Symbolic regression appears to be quite unknown to the machine learning community as only a few works have been reported on the use of SR for control of dynamic systems. For instance, modeling of the value function by means of genetic programming is presented in \cite{onderwater2016tsmc}, where analytic descriptions of the value function are obtained based on data sampled from the optimal value function. Another example is the work \cite{alibekov2016cdc}, where SR is used to construct an analytic function, which serves as a proxy to the value function and a continuous policy can be derived from it. A multi-objective evolutionary algorithm was proposed in \cite{branke2015tec}, which is based on interactive learning of the value function through inputs from the user. SR is employed to construct a smooth analytic approximation of the policy in \cite{kubalik2017ifac}, using the data sampled from the interpolated policy.

To our best knowledge, there have been no reports in the literature on the use of symbolic regression for constructing a process model in model-based control methods. We argue that the use of SR for model learning is a valuable element missing from the current nonlinear control schemes and we demonstrate its usefulness.

In this paper, we extend our previous work \cite{Derner18-ICRA,Derner18-IROS}, which indicated that SR is a suitable tool for this task. It does not require any basis functions defined a priori and contrary to (deep) neural networks it learns accurate, parsimonious models even from very small data sets. Symbolic regression can handle also high-dimensional problems and it does not suffer from the exponential growth of the computational complexity with the dimensionality of the problem, which we demonstrate on an enriched set of experiments including a complex bipedal walking robot system. In this work, we extend the use of the method to the class of input--output models, which are suitable in cases when the full state vector cannot be measured. By testing our method with two different state-of-the-art genetic programming algorithms, we demonstrate that the method is not dependent on the particular choice of the SR algorithm.

The paper is organized as follows. Sections~\ref{sec:theory} and~\ref{sec:method} present the relevant context for model learning and the proposed method. The experimental evaluation of the method is reported in Section~\ref{sec:experiments} and the conclusions are drawn in Section~\ref{sec:conclusions}. \ref{sec:rl} describes the RL method used in this paper and \ref{sec:tables} lists detailed results of the experiments.

\section{Theoretical Background}
\label{sec:theory}

The discrete-time nonlinear state-space process model is described as
\begin{equation}
\begin{aligned}
x_{k+1} &= f(x_k,u_k)
\end{aligned}
\label{eq:model}
\end{equation}
with the state $x_k, x_{k+1} \in \mathcal{X} \subset \mathbb{R}^n$ and the input $u_k \in \mathcal{U} \subset \mathbb{R}^m$. Note that the actual process can be stochastic (typically when the sensor readings are corrupted by noise), but in this paper we aim at constructing a deterministic process model \eqref{eq:model}.

The full state vector cannot be directly measured for a vast majority of processes and a state estimator would have to be used. In the absence of an accurate process model, such a reconstruction is inaccurate and has a negative effect on the overall performance of the control algorithm on the real system. Note that this problem has not been explicitly addressed in the literature, as most results are demonstrated on simulation examples in which the state information is available.

Therefore, next to state-space models, we also investigate the use of dynamic input--output models of the NARX (nonlinear autoregressive with exogenous input) type. The NARX model establishes a relation between the past input--output data and the predicted output:
\begin{equation}
y_{k+1}=g\left(y_k,y_{k-1},\ldots,y_{k-n_y+1},u_k,u_{k-1},\ldots,u_{k-n_u+1}\right),
\label{eq:model0-narx}
\end{equation}
where $n_y$ and $n_u$ are user-defined integer parameters based on the expected system's order, and $g$ is a static function, different from the function $f$ used in the state-space model \eqref{eq:model}.

For the ease of notation, we group the lagged outputs and inputs into one vector:
$$
 \varphi_k = [y_k,y_{k-1}\ldots,y_{k-n_y+1},u_{k-1},\ldots,u_{k-n_u+1}]
$$
and write model \eqref{eq:model0-narx} as:
\begin{equation}
y_{k+1}=g\left(\varphi_k, u_k\right) .
\label{eq:model-narx}
\end{equation}

\begin{figure}[htbp]
  \centering
  \includegraphics[width=0.6\columnwidth]{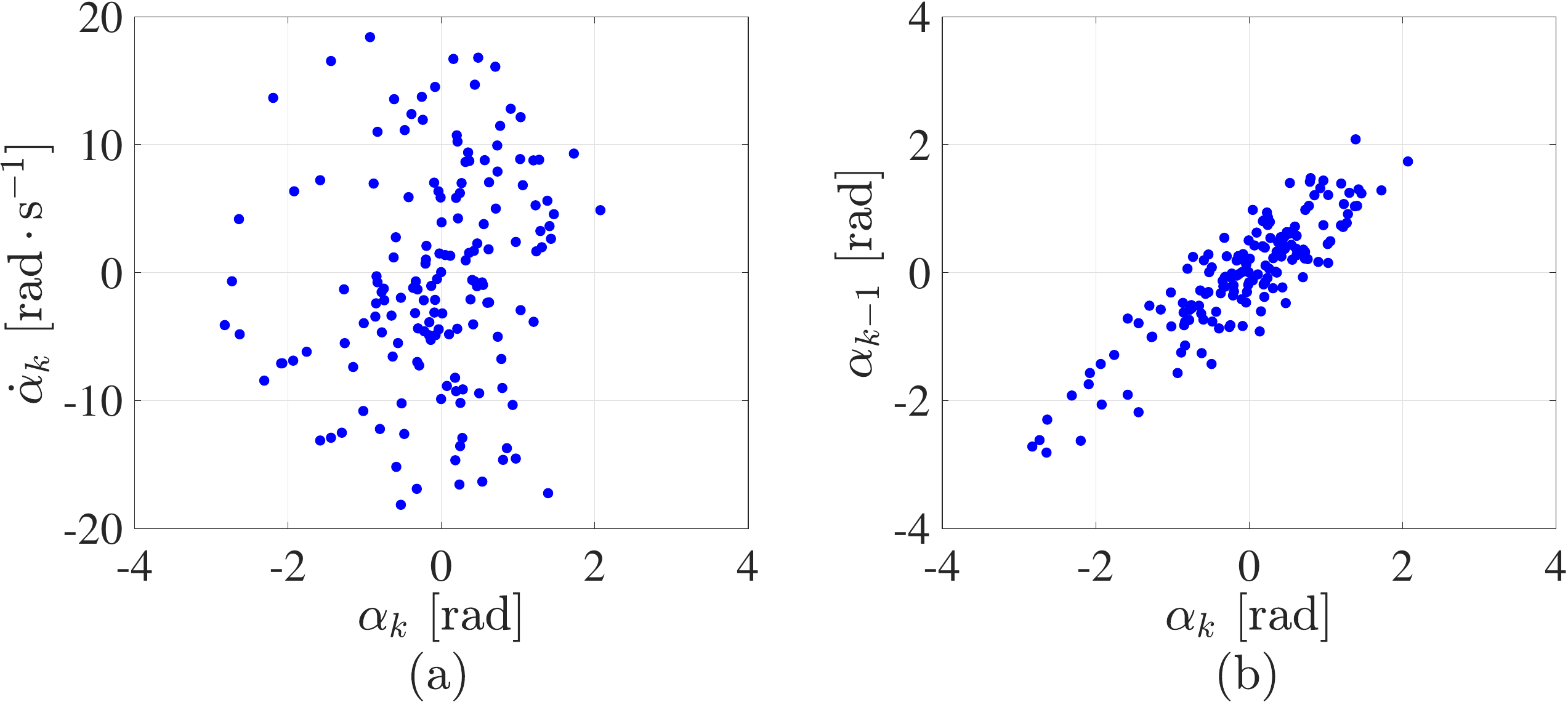}
  \caption{An example of trajectory samples obtained from the real inverted pendulum (see Section~\ref{sec:inverted-pendulum}) in the original state space (a), and in the space formed by the current and previous output (b).}
  \label{fig:compare-pend-scatter}%
\end{figure}

Note that in this setting, the model function and also the control policy are found from data samples which live in a space that is very different from the state space. The lagged outputs $y_k, y_{k-1}, \ldots, y_{k-n_y+1}$ are highly correlated and therefore span a deformed space. This presents a problem for many types of approximators. For instance, basis functions defined by the Cartesian product of the individual lagged variables will cover the whole product space $y_k \times y_{k-1} \times \ldots \times y_{k-n_y+1}$, while data samples only span a small, diagonally oriented part of the space, as illustrated in Figure~\ref{fig:compare-pend-scatter}. The SR approach described in this paper does not suffer from such drawbacks.

In this paper, we use reinforcement learning as the control method of choice. Please refer to \ref{sec:rl} for details on the RL method used.

\section{Method\label{sec:method}}

In this section, we explain the principle of our method, briefly describe two variants of genetic programming algorithms used in this work, and discuss the computational complexity of our approach.

\subsection{Symbolic Regression}
\label{sec:sr}

Symbolic regression is employed to approximate the unknown state transition function $f$ in the state-space model \eqref{eq:model} or $g$ in the input--output model \eqref{eq:model0-narx}. The analytic expressions describing the process to be controlled are constructed through genetic programming. SR methods were reported in the literature to work faster when using a linear combination of evolved nonlinear functions instead of evolving the whole analytic expression at once \cite{Arnaldo2014,Arnaldo2015}. Therefore, we define the class of analytic state-space models as:
\begin{equation}
    f(x,u) = \beta_0 + \sum_{i=1}^{n_f} \beta_i f_i(x,u)
\label{eq:sr-stsp}
\end{equation}
and the class of analytic input--output (NARX) models as:
\begin{equation}
    g(\varphi,u) = \beta_0 + \sum_{i=1}^{n_f} \beta_i g_i(\varphi,u)\,.
\label{eq:sr-narx}
\end{equation}
The nonlinear functions $f_i(x,u)$ or $g_i(\varphi,u)$, called features, are constructed from a set of user-defined elementary functions. These functions can be nested and are evolved by means of standard evolutionary algorithm operations, such as mutation, so that the mean-square error calculated over the training data set is minimized. No a priori knowledge on the structure of the nonlinear model is needed. The set of elementary functions may be broad to let the SR algorithm select functions that are most suitable for fitting the given data. However, it is also possible to provide the algorithm with a partial knowledge about the problem. A narrower selection of elementary functions restricts the search space and speeds up the evolution process.

To avoid over-fitting, we control the complexity of the regression model by imposing a limit on the number of features $n_f$ and the maximum depth $d$ of the tree representation of the features. The coefficients $\beta_i$ are estimated by least squares.

\subsection{Genetic Programming Methods Used}
\label{sec:gp_methods}

In order to demonstrate that our method is not dependent on the particular choice of the SR algorithm, we test our approach with two different genetic programming methods: a modified version of Single Node Genetic Programming (SNGP) \cite{Jackson2012a,Jackson2012b,Kubalik17-IJCCI} and a modified version of Multi-Gene Genetic Programming (MGGP) \cite{Zegklitz17}. Both methods have been successfully used for symbolic regression, with several applications in the RL and robotics domains \cite{kubalik19arxiv,kubalik2017ifac,Derner18-IROS,Derner18-ICRA}.

SNGP is a graph-based genetic programming technique that evolves a population of nodes organized in the form of an ordered linear array. The nodes can be of various types depending on the particular problem. In the context of SR, the node can either be a terminal, i.e., a constant or a variable, or some operator or function chosen from a set of functions defined by the user for the problem at hand. The individuals are interconnected in the left-to-right manner, meaning that an individual can act as an input operand only of those individuals which are positioned to its right in the population. Thus, the whole population represents a graph structure with multiple expressions rooted in the individual nodes. Expressions rooted in the function nodes can represent non-linear symbolic functions of various complexity. The population is evolved through a local search procedure using a single reversible mutation operator.

MGGP is a tree-based genetic programming algorithm utilizing multiple linear regression. The main idea behind MGGP is that each individual is composed of multiple independent expression trees, called genes, which are put together by a linear combination to form a single final expression. The parameters of this top-level linear combination are computed using multiple linear regression where each gene acts as an independent feature. In this article, we build upon a particular implementation of MGGP -- GPTIPS2 \cite{Searson2015}. This particular instance of MGGP uses two crossover operators: (i) high-level crossover that combines gene sets of two parents; (ii) low-level crossover which is a classical Koza-style \cite{koza1992book} subtree crossover operating on corresponding pairs of parental genes. Also, there are two mutation operators: (i) subtree mutation, which is a classical Koza-style subtree mutation; (ii) constant mutation, which alters the numerical values of leaves representing constants. Both the crossover and mutation operators are chosen stochastically.

Detailed explanation of these algorithms and their parameters is beyond the scope of this paper and we refer the interested reader to \cite{Kubalik17-IJCCI} and \cite{Zegklitz17}.

\subsection{Computational Complexity}

The computational complexity of the symbolic regression algorithms used in this work increases linearly with the size of the training data set as well as with the dimensionality of the problem. For example, considering a problem with one-dimensional state, one-dimensional input, one-dimensional output, and a data set of 1000 samples, a single run of the SNGP or MGGP algorithm with the default configuration takes about 3 minutes on a single core of a standard desktop PC. For a system with a 14-dimensional regressor and a 6-dimensional input, a single run takes up to 20 minutes.

\section{Experimental Results\label{sec:experiments}}

We have carried out experiments with three nonlinear systems: a mobile robot, a 1-DOF inverted pendulum and a bipedal walking robot. The data, the codes and the detailed configuration of the experiments is available in our repository\footnote{\url{https://github.com/erik-derner/symbolic-regression}}.

The simulation experiment with the mobile robot illustrates the use of the presented method, showing the precision and compactness of the models found in the case where the ground truth is known (Section~\ref{sec:mobile-robot}). We show that the method is not dependent on the particular choice of the SR algorithm by comparing the performance of two SR methods, SNGP and MGGP. The subsequent experiment with the walking robot presents a more complex example and shows the performance of the method in a high-dimensional space (Section~\ref{sec:walking-robot}). With this example, we demonstrate the ability of the method to construct standard state-space models as well as input--output (NARX) models and we show how the method performs compared to two deep neural networks with different architectures. We conclude our set of experiments with the inverted pendulum system (Section~\ref{sec:inverted-pendulum}). Similarly as in the experiment with the mobile robot, we evaluate the method with SNGP and MGGP, and we compare the results to two alternative approaches: neural networks and local linear regression. In addition to measuring the model prediction error, we perform real-time closed-loop control experiments with a lab setup to evaluate the performance of the algorithm in real-world conditions.

%%%%%%%%%%%%%%%%%%%%%%%%%%%%%%%%%%%%
% MOBILE ROBOT
%%%%%%%%%%%%%%%%%%%%%%%%%%%%%%%%%%%%

\subsection{Mobile Robot}
\label{sec:mobile-robot}

\begin{figure}[htbp]
\centerline{
\subfigure[]{\includegraphics[width=3.6cm]{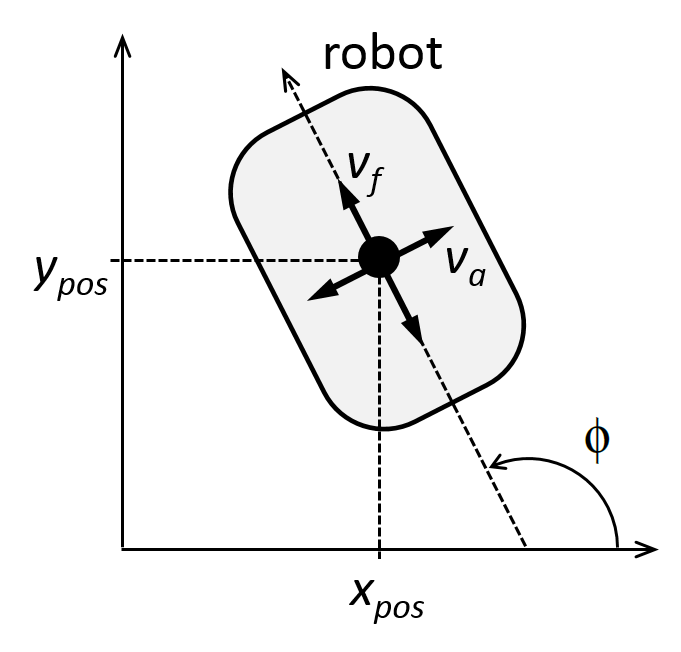}} \qquad
\subfigure[]{\includegraphics[width=3.1cm]{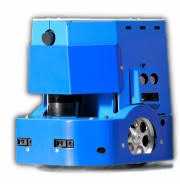}}
}
    \caption{Mobile robot schematic (a) and photograph (b).}
    \label{fig:smr}%
\end{figure}

The state of a two-wheel mobile robot, see Figure~\ref{fig:smr} and \cite{faigl2010syrotek}, is described by $x = [x_{pos}, y_{pos}, \phi]^\top$, with $x_{pos}$ and $y_{pos}$ the position coordinates and $\phi$ the heading. The control input is $u = [v_f, v_a]^\top$, where $v_f$ represents the forward velocity and $v_a$ the angular velocity of the robot.

The continuous-time dynamic model of the robot is:

\begin{align}
\label{eq:smr}
\begin{split}
\dot x_{pos} &= v_f \, \cos(\phi) , \\
\dot y_{pos} &= v_f \, \sin(\phi) , \\
\dot \phi &= v_a\,.
\end{split}
\end{align}

\subsubsection{Data Sets\label{sec:mobile-robot-datasets}}

We generated a noise-free data set by using the Euler method to simulate the differential equations \eqref{eq:smr}. With a sampling period $T_s = 0.05$\,s, the discrete-time approximation of \eqref{eq:smr} becomes:
\begin{align}
\label{eq:tf-mobile-robot}
\begin{split}
x_{pos, \, k+1} &= x_{pos, \, k} + 0.05 \, v_{f, \, k} \, \cos(\phi) , \\
y_{pos, \, k+1} &= y_{pos, \, k} + 0.05 \, v_{f, \, k} \, \sin(\phi) , \\
\phi_{k+1} &= \phi_{k} + 0.05 \, v_{a, \, k}\,.
\end{split}
\end{align}

We generated training data sets of different sizes $n_s$. The initial state $x_0$ and the control input $u_k$ for the whole simulation were randomly chosen from the ranges:
\begin{align}
\label{eq:mobile-robot-limits}
\begin{split}
x_{pos} &\in [-1, 1] \, \mathrm{m} , \\
y_{pos} &\in [-1, 1] \, \mathrm{m} , \\
\phi &\in [-\pi, \pi] \, \mathrm{rad} , \\
v_f &\in [-1, 1] \, \mathrm{m \cdot s}^{-1} , \\
v_a &\in \left[-{\pi\over 2}, {\pi\over 2}\right] \, \mathrm{rad \cdot s}^{-1} .
\end{split}
\end{align}

A test data set was generated in order to assess the quality of the analytic models on data different from the training set. The test data set entries were sampled on a regular grid with 11 points spanning evenly each state and action component domain, as defined by (\ref{eq:mobile-robot-limits}). These samples were stored together with the next states calculated by using the Euler approximation.

\subsubsection{Experiment Setup}

The purpose of this experiment was to test the ability of the SNGP and MGGP algorithms to recover from the data the analytic process model described by a known state-transition function. In order to assess the performance depending on the size of the data set and the complexity of the model, different combinations of the number of features $n_f$ and the size of the training set $n_s$ were tested. As the used algorithms only allow modeling one output at a time, they were run independently for each of the state components $x_{pos, \, k+1}$, $y_{pos, \, k+1}$ and $\phi_{k+1}$.

The size of the SNGP population was set to 500 individuals and the evolution duration to 30000 generations. The set of elementary functions was defined as $\{*, \, +, \, -, \, \mathrm{sin}, \, \mathrm{cos}\}$. The maximum depth $d$ of the evolved nonlinear functions was set to 7 and the number of features was $n_f \in \{1, 2, 10\}$. To ensure a fair evaluation, the parameters of the MGGP algorithm were set similarly to provide both methods with a comparable amount of computational resoruces, taking into account the conceptual differences between the two algorithms.

\subsubsection{Results}

The models found by symbolic regression were evaluated by calculating the RMSE median on the test data set over 30 independent runs of the SR algorithm. Note that each run yields a different model because the evolution process is guided by a unique sequence of random numbers. The results are listed in Table~\ref{tab:mobile-robot} in \ref{sec:tables}.

An example of a process model found by running SNGP with the parameters $n_f=2$ and $n_s=100$ is:
\begin{align}
\label{eq:tf-mobile-robot-symbolic}
\begin{split}
\hat{x}_{pos, \, k+1} =& \; 1.0 \, x_{pos, \, k} + 0.0499998879 \, v_{f, \, k} \, \cos(\phi_k) \, , \\
\hat{y}_{pos, \, k+1} =& \; 1.000000023 \, y_{pos, \, k} + 0.0500000056 \, v_{f, \, k} \sin(\phi_k) + 0.0000000191 \, , \\
\hat{\phi}_{k+1} =& \; 0.9999982931 \, \phi_k + 0.0500000536 \, v_{a, \, k} - 0.0000059844 \, .
\end{split}
\end{align}
The coefficients are rounded to 10 decimal digits in order to demonstrate the magnitude of the error compared to the original Euler approximation (\ref{eq:tf-mobile-robot}). The results show that even with a small training data set, a precise, parsimonious analytic process model can be found based on noise-free data.

The results also demonstrate how the number of features $n_f$ plays an important role in the setting of the experiment parameters. In general, the RMSE decreases with an increasing number of features, whereas the complexity naturally grows by adding more features to the final model~\eqref{eq:sr-stsp}. The higher RMSE error when using only one feature is caused mainly by the fact that all parameters have to be evolved by the genetic algorithm, which is hard. On the other hand, when using more features, the least squares method can quickly and accurately find the coefficients of the features. These results support our choice to define the class of analytic models as a linear combination of features, as explained in Section~\ref{sec:sr}. As a corollary, if the outline of the model structure is known in advance, it is recommended to set the number of features at least equal to the number of terms expected in the underlying function. Otherwise, it is advisable to set the number of features large enough, e.g. $n_f = 10$.

%%%%%%%%%%%%%%%%%%%%%%%%%%%%%%%%%%%%
% WALKING ROBOT
%%%%%%%%%%%%%%%%%%%%%%%%%%%%%%%%%%%%

\subsection{Walking Robot}
\label{sec:walking-robot}

\begin{figure}[htbp]
\centering
\subfigure[]{
\includegraphics[width=2.5cm]{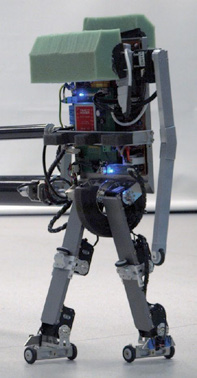}}\qquad
\subfigure[]{
\includegraphics[width=2.39cm]{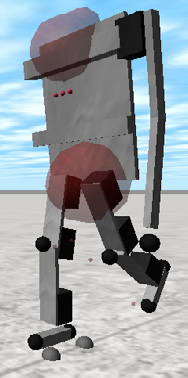}}\qquad
\caption{The walking robot LEO: photograph (a) and simulation model rendering (b).
\cite{Koryakovskiy2017}.}
\label{fig:LEO}
\end{figure}

The robot LEO is a 2D bipedal walking robot \cite{Schuitema2010}, see Figure~\ref{fig:LEO}. It has 7 actuators: two in the ankles, knees and hips and one in its shoulder that allows the robot to stand up after a fall. LEO is connected to a boom with a parallelogram construction. This keeps the hip axis always horizontal, which makes it effectively a 2D robot and thus eliminates the sideways stability problem.

The state vector of LEO $x = [{\psi}, \dot{\psi}]^\top$ consists of 14 components, where
\begin{equation}
\psi = [\psi_{TRS}, \psi_{LH}, \psi_{RH}, \psi_{LK}, \psi_{RK}, \psi_{LA}, \psi_{RA}]^\top
\end{equation}
represents the angles of the torso, left and right hip, the knee and the ankle. Likewise,
\begin{equation}
\dot{\psi} = [\dot{\psi}_{TRS}, \dot{\psi}_{LH}, \dot{\psi}_{RH}, \dot{\psi}_{LK}, \dot{\psi}_{RK}, \dot{\psi}_{LA}, \dot{\psi}_{RA}]^\top
\end{equation}
are the angular velocities of the torso, hips, knees and ankles. The action space of LEO comprises the voltage inputs to the seven joint actuators.

\subsubsection{Data Sets}

In order to apply symbolic regression, the walking robot LEO was modeled using the Rigid Body Dynamics Library (RBDL) \cite{Felis2016} and the data sets were generated using the Generic Reinforcement Learning Library (GRL) \cite{Caarls2018}, which  allowed us to record trajectories while the robot was learning to walk.

We split the data set into two disjoint subsets: a training set and a test set. Both subsets are composed of consecutive samples from the simulation, which was run with a sampling period $T_s = 0.03$\,s.

\subsubsection{Experiment Setup}

The experiment was designed to evaluate the performance of our method on a more complex, high-dimensional example and to construct input--output (NARX) models in addition to the standard state-space models. We chose to use only the SNGP algorithm for this experiment and the main parameters were configured as follows. The population size was set to 500 and the number of generations to 30000. The depth limit $d$ was fixed to 5 and the number of features was $n_f \in \{1, 5, 10\}$. The elementary function set was defined as $\{*, +, -, \sin, \cos, \mathrm{square}, \mathrm{cube}\}$.

During the simulation used to obtain the data sets, the shoulder was not actuated. Therefore, the input vector had only 6 components, one for each actuator. As in the case of the mobile robot, SNGP was run separately for each of the 14 state components.

In \emph{Experiment~B1}, we used SR to generate standard state-space models. In \emph{Experiment~B2}, we generated input--output models with the regression vector defined as $\varphi = [\psi_k, \psi_{k-1}, u_{k-1}]$.

\subsubsection{Results}

In order to evaluate the ability of SNGP to approximate the state-transition function, we calculated the RMSE medians over 30 runs of the algorithm on the test data set. The results for the state-space models are reported in Table~\ref{tab:walking-robot-stsp} and for the input--output models in Table~\ref{tab:walking-robot-narx} in \ref{sec:tables}.

The results show the expected trend, which can be seen in all experiments: the quality of the models improves with the size of the training data set. However, it is noteworthy that the difference between the RMSE for models trained on 100 samples and for those trained on 5000 samples are in most cases negligible. This confirms our earlier observation that SR can be used to find accurate analytic process models on batches of data as small as 100 samples~\cite{Derner18-ICRA} even for high-dimensional systems.

The results for the input--output models are generally just slightly worse than those for the state-space models, with the benefit of speeding up the algorithm by reducing the number of modeled variables to a half.

\subsubsection{Comparison with Alternative Methods}
\label{sec:walking-robot-alternative-methods}

Deep neural networks are widely used to model an unknown system. In order to compare our method to alternative state-of-the-art methods, we have constructed two different neural networks:
\begin{itemize}
 \item Deep neural network DNN-A was implemented in PyTorch. It consists of an input linear layer of size $20 \times 200$, followed by three linear layers with the size of $200 \times 200$, with a ReLU activation function used after each linear layer. The output layer has $200 \times 14$ units. The batch size was set to 32. The SGD algorithm \cite{Kiefer1952} was used with the learning rate of $8.5 \times 10^{-4}$.
 \item Deep neural network DNN-B was implemented in TensorFlow. It is a fully connected network with 1 hidden layer, consisting of 512 units with ELU nonlinearity and 50\% dropout. The batch size was set to 8. The Adam optimizer \cite{Kingma2014} was used with a learning rate of $10^{-3}$ and early stopping.
\end{itemize}

\begin{figure}[htbp]
  \centering
  \includegraphics[width=0.47\columnwidth]{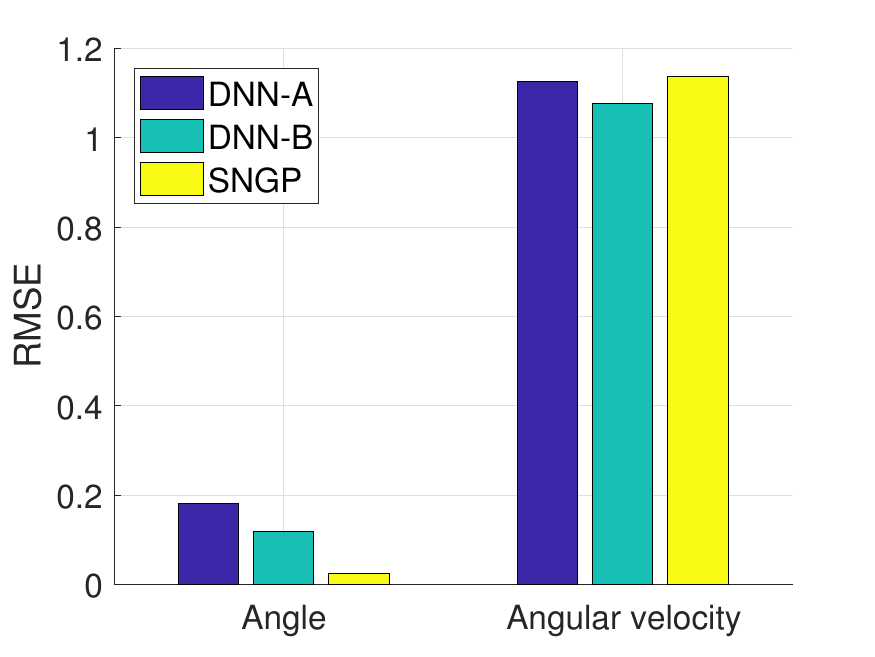}
  \caption{Comparison of two DNN variants with the SNGP algorithm on the walking robot example. The bars show the mean RMSE over the 7 angles and over the 7 angular velocities on the test data set.}
  \label{fig:leo-dnn-comparison}
\end{figure}

We chose the RMSE medians of SNGP with $n_s = 1000$ and $n_f = 10$ as the benchmark configuration. State-space models were used in this scenario. The training and test sets were the same for all compared methods. Figure~\ref{fig:leo-dnn-comparison} shows an overview of the performance of the two variants of DNN compared to the SNGP algorithm and detailed results are presented in Table~\ref{tab:walking-robot-dnn} in \ref{sec:tables}. The results show that the SNGP algorithm is able to find substantially better models than the neural networks for the angles, while the performance on the angular velocities is comparable among all the tested methods.

%%%%%%%%%%%%%%%%%%%%%%%%%%%%%%%%%%%%
% INVERTED PENDULUM
%%%%%%%%%%%%%%%%%%%%%%%%%%%%%%%%%%%%

\subsection{Inverted Pendulum}
\label{sec:inverted-pendulum}

\begin{figure}[htbp]
\centerline{
\subfigure[]{\includegraphics[width=3cm]{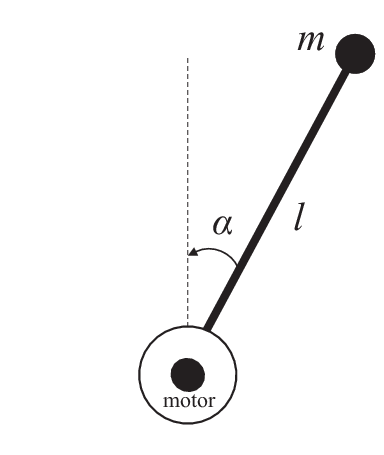}} \qquad
\subfigure[]{\includegraphics[width=4.3cm]{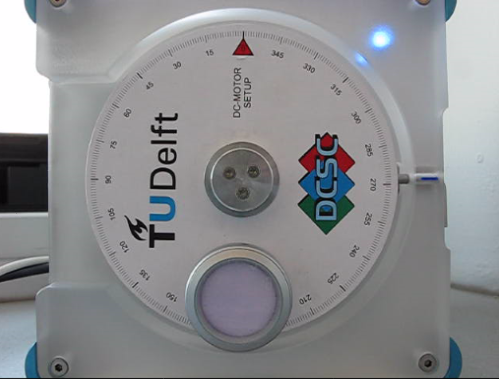}}
}
    \caption{Inverted pendulum schematic (a) and the real inverted pendulum system (b).}
    \label{fig:ips}%
\end{figure}

The inverted pendulum system consists of a weight of mass $m$ attached to an actuated link which rotates in the vertical plane, see Figure~\ref{fig:ips}a. The state vector is $x = [\alpha, \dot\alpha]^\top$, where $\alpha$ is the angle and $\dot\alpha$ is the angular velocity of the link. The control input is the voltage $u$. The continuous-time model of the pendulum dynamics is:
\begin{equation}\label{eq:ips}
\ddot\alpha = \frac{1}{J} \cdot \left(\frac{K}{R} \, u - m \, g \, l \, \sin(\alpha) - b \, \dot\alpha - \frac{K^2}{R} \, \dot\alpha - c \, \mathrm{sign}(\dot\alpha) \right)
\end{equation}
with $J = 1.7937 \times 10^{-4} \, \mathrm{kg \cdot m}^2$, $m = 0.055 \, \mathrm{kg}$, $g = 9.81 \, \mathrm{m \cdot s}^{-2}$, $l = 0.042 \, \mathrm{m}$, $b = 1.94 \times 10^{-5} \, \mathrm{N \cdot m \cdot s \cdot rad}^{-1}$, $K = 0.0536 \, \mathrm{N \cdot m \cdot A}^{-1}$, $R = 9.5 \, \Omega$ and $c = 8.5 \times 10^{-4} \, \mathrm{kg \cdot m}^{2} \mathrm{\cdot s}^{-2}$. The angle is $\alpha = 0$ or $\alpha = 2\pi$ for the pendulum pointing down and $\alpha = \pi$ for the pendulum pointing up.

The reward function used in the RL experiments was defined as follows:
\begin{equation}
\begin{aligned}
&\rho(x_k, u_k, x_{k+1}) = - 0.5 |\alpha_r - \alpha_k| - 0.01 |\dot\alpha_r - \dot\alpha_k| - 0.05 |u_k| ,
\end{aligned}
\label{eq:reward-pendulum}
\end{equation}
where $[\alpha_r, \dot\alpha_r]^\top$ is a constant reference (goal) state.

\subsubsection{Data Collection}

As we will present an experiment with the inverted pendulum performing a control task, we start this section by a short overview of the data collection methods used. Two different situations can be distinguished: initial model learning and model learning under a given policy.

\paragraph{Initial Model Learning}

At the beginning, when the control policy is not yet available, the system can be excited by a test signal in order to obtain a sufficiently rich data set. Various methods for designing suitable test signals are described in the literature, such as the generalized binary noise (GBN) sequence \cite{Tulleken90}. The important parameters to be selected are the input signal amplitude, the way the random signal is generated (e.g., the `switching' probability) and the experiment duration.

\paragraph{Model Learning Under a Given Policy}

Once an acceptable control policy has been learned, the system can be controlled to execute the required task. Data can be collected while performing the control task and used to further improve the model. As the information captured in the data under steady operating conditions might not be sufficient in certain situations, the control input can be adjusted by adding a test signal in this case as well. The characteristics of this test signal are usually different from the one used for initial model learning; for instance, it typically has a lower amplitude.

\subsubsection{Data Sets}
\label{sec:swing-up-data-sets}

We used both simulated and real measured data in the experiments with the inverted pendulum. In all experiments, the discrete-time sampling period used was $T_s = 0.05$\,s.

At first, we generated a noise-free data set for \emph{Experiment~C1} by using the Euler method to simulate the differential equation (\ref{eq:ips}):
\begin{align}
\label{eq:tf-pend}
\begin{split}
\alpha_{k+1} =& \; \alpha_{k} + 0.05 \, \dot\alpha_{k} , \\
\dot\alpha_{k+1} =& \; 0.9102924564 \, \dot\alpha_{k} - 0.2369404025 \, \mathrm{sign}(\dot\alpha_{k}) \\
& + 1.5727561084 \, u_k - 6.3168590065 \, \sin(\alpha_k) .
\end{split}
\end{align}

The data set for \emph{Experiment~C2} was created by integrating (\ref{eq:ips}) by using the fourth-order Runge-Kutta method and adding Gaussian noise. The transformation from the original states $x = [\alpha, \dot\alpha]^\top$ to the states with Gaussian noise $x_n = [\alpha_n, \dot\alpha_n]^\top$ is defined as
\begin{align}
\label{eq:noise}
\begin{split}
\alpha_n &= \alpha + \pi \lambda r_{n,1}\,, \\
\dot\alpha_n &= \dot\alpha + 40 \lambda r_{n,2}\,,
\end{split}
\end{align}
where $r_{n,1}, \, r_{n,2}$ are random numbers drawn from a normal distribution with zero mean and a standard deviation of 1. The constant $\lambda \in \{ 0, 0.01, 0.05, 0.1 \}$ controls the amount of noise and the constants $\pi$ and $40$ make sure that the added noise is approximately proportional to the range of each variable.

In both \emph{Experiments~C1} and \emph{C2}, the initial state was $\alpha = 0$, $\dot\alpha = 0$ and the control input was chosen randomly at each time step $k$ from the range $u_k \in [-5, 5] \, \mathrm{V}$.

The test data sets were created similarly as in Section~\ref{sec:mobile-robot-datasets}. The samples were generated on a regular grid of $31 \times 31 \times 31$ points, spanning the state and action domain: $\alpha \in [-\pi, \pi] \, \mathrm{rad}$, $\dot\alpha \in [-40, 40] \, \mathrm{rad \cdot s}^{-1}$ and $u \in [-5, 5] \, \mathrm{V}$. For all samples, the next states in the test set for \emph{Experiment~C1} were calculated using the Euler approximation. In \emph{Experiment~C2}, we generated a noise-free test set by applying the fourth-order Runge-Kutta method to all samples on the grid.

\begin{figure}[htbp]
  \centering
  \includegraphics[width=0.6\columnwidth]{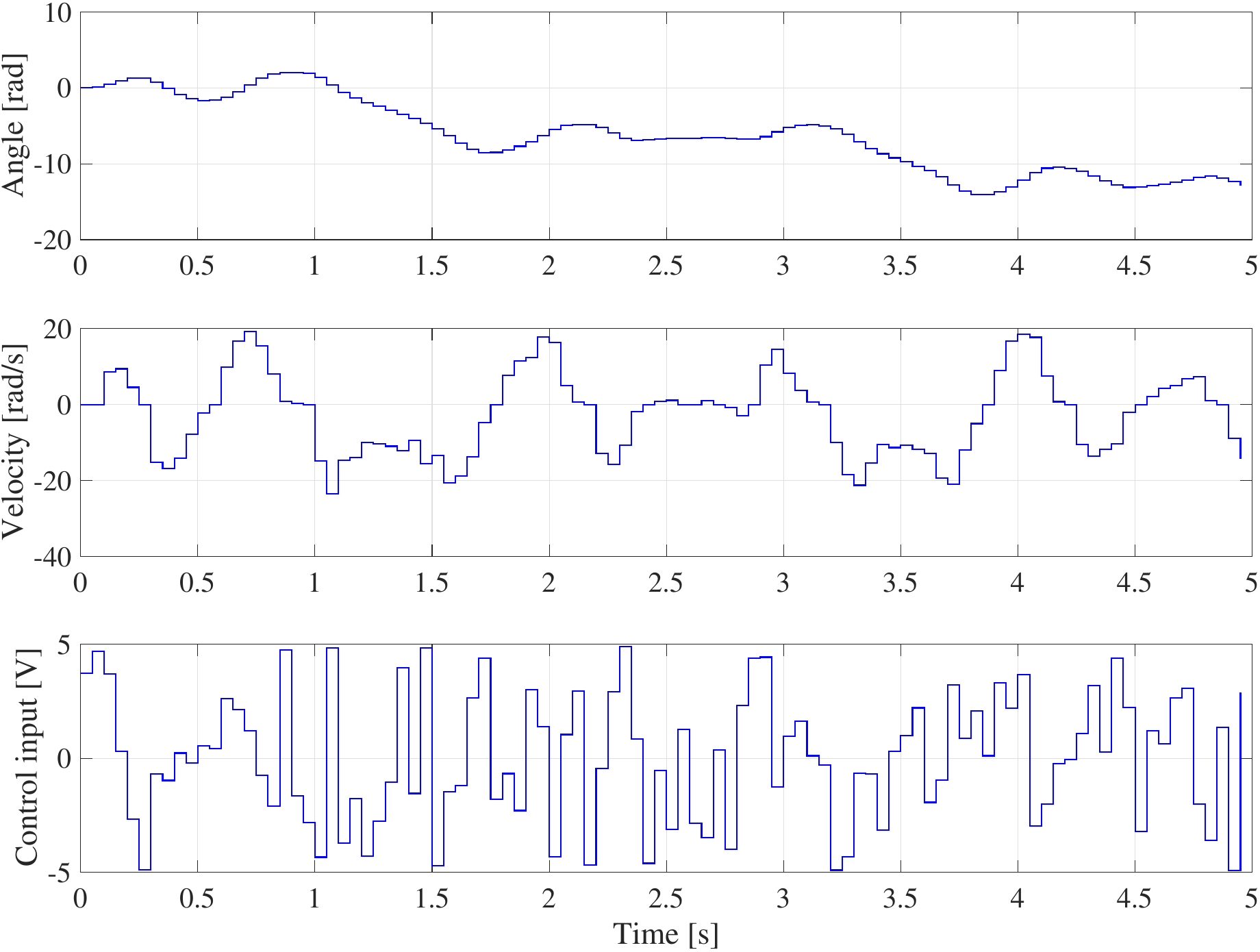}
  \caption{Initial data set obtained on the real inverted pendulum system as a response to the random input shown in the bottom panel.}
  \label{fig:pendulum-random-data}
\end{figure}

The real data for \emph{Experiment~C3} were measured on the real inverted pendulum system shown in Figure~\ref{fig:ips}b. At first, the system was excited by applying a uniformly distributed random control input $u_k$ within the range $[-5, 5] \, \mathrm{V}$ at each time step $k$. The random interaction with the system lasted for 5~seconds and the recorded data set comprised 100 samples. The data are shown in Figure~\ref{fig:pendulum-random-data}. The data set was later enriched by samples recorded while applying the control policy \eqref{eq:hillclimb} to perform the swing-up task on the real system, which will be described in the following section.

The sequences recorded for \emph{Experiment~C3} were split into training and test subsets. Every third sample was used for the test set, while the remaining samples formed the training set. In all experiments, the reported RMSE values were calculated on the respective test data set.

\subsubsection{Experiment Setup}

Similarly as in the experiment with the mobile robot, the SNGP and MGGP algorithms were first employed in \emph{Experiment~C1} to test the ability of SR to generate precise models for the inverted pendulum system using a data set generated by the Euler method. The experiment serves to evaluate how the training data set size $n_s$ and the number of features $n_f$ influence the quality of the model.

\emph{Experiment~C2} demonstrates how the analytic process models are evolved using the Runge-Kutta simulation data set with noise. The maximum number of features $n_f$ in the symbolic regression algorithms was set to 10 in order to facilitate the evolution of models capturing the more complex underlying function. This experiment tests the behavior of the method in environments with noisy measurements.

We conclude the experiments with \emph{Experiment~C3}, which shows the intended use of the method within RL on the example of the underactuated swing-up task, performed on a real inverted pendulum system. The control goal is to stabilize the pendulum in the unstable equilibrium $x_r = [\alpha_r, \dot\alpha_r]^\top = [\pi, 0]^\top$. As the input is limited to the range $u \in [-2, 2]$\,V, the available torque is insufficient to push the pendulum directly up from the majority of initial states, and therefore it has to be first swung back and forth to gather energy. At first, we constructed 30 analytic models using the data set recorded under random input and then selected the model with the lowest RMSE on the test set. This \emph{initial model} was employed to calculate the policy for the swing-up task (see \ref{sec:rl}). To find an approximation of the optimal value function, we used the fuzzy V-iteration algorithm \cite{Busoniu10-Automatica}. We applied the policy to the real system in four independent runs, starting at the initial state $x_0 = [0, 0]^\top$. In addition, we performed other four swing-ups with exploration noise added to the control input. The exploration noise was normally distributed with the standard deviation ranging from 0.2 to 0.5\,V. All eight sequences, each consisting of approximately 50 measurements, were added to the initial data set recorded under random input. Using this extended data set, 30 refined analytic process models were learned and the model with the lowest error on the test set was chosen as the final \emph{refined model}. Like in \emph{Experiment~C2}, the number of features was set to $n_f = 10$ to facilitate modeling the more complex state-transition function.

In all experiments, the size of the SNGP population was set to 500 and the evolution was limited to 30000 generations. The elementary function set was $\{*, \, +, \, -, \, \mathrm{sin}, \, \mathrm{cos}, \, \mathrm{sign}\}$. The maximum depth $d$ was set to~7. In \emph{Experiment~C1}, various numbers of features were tested: $n_f \in \{1, 2, 10\}$ for $\alpha$ and $n_f \in \{1, 4, 10\}$ for $\dot\alpha$.  The parameters of the MGGP algorithm in \emph{Experiment~C1} and \emph{C2} were set similarly, taking into account the conceptual differences between the two algorithms to allow for a fair comparison.

\subsubsection{Results}

The results of \emph{Experiment~C1} are summarized in Table~\ref{tab:e2} in \ref{sec:tables} for the SNGP and MGGP algorithm. Similarly as in the previous examples, the results indicate that the precision of the models increases with increasing number of features. The overall performance of both SR algorithms is comparable.

An example of an analytic process model found with the parameters $n_f=2$ for $\alpha$, $n_f=4$ for $\dot\alpha$ and $n_s=20$ is:
\begin{align}
\label{eq:tf-pend-symbolic}
\begin{split}
\hat\alpha_{k+1} =& \; \alpha_{k} + 0.05 \, \dot\alpha_{k} - 0.0000000001 \, , \\
\hat{\dot\alpha}_{k+1} =& \; 0.9102924745 \, \dot\alpha_{k} - 0.2369403835 \, \mathrm{sign}(\dot\alpha_{k})+ 1.5727561072 \, u_k \\
 -& \; 6.3168589936 \, \sin(\alpha_k) + 0.0000000013 \, .
\end{split}
\end{align}

The error of the analytic model w.r.t. the Euler approximation (\ref{eq:tf-pend}) is very small. These results confirm that the proposed method can find precise models even on small data sets.

The results of \emph{Experiment~C2} presented in Table~\ref{tab:e3} in \ref{sec:tables} show that the analytic models are able to approximate the state-transition function well even on data with a reasonable amount of noise. The use of the Runge-Kutta method to generate data sets leads to substantially more complicated models than when using the data generated by using the Euler method. Again, the performance of the SNGP and MGGP algorithm is comparable.

\begin{figure}[htbp]
\subfigure[]{\includegraphics[width=0.47\columnwidth]{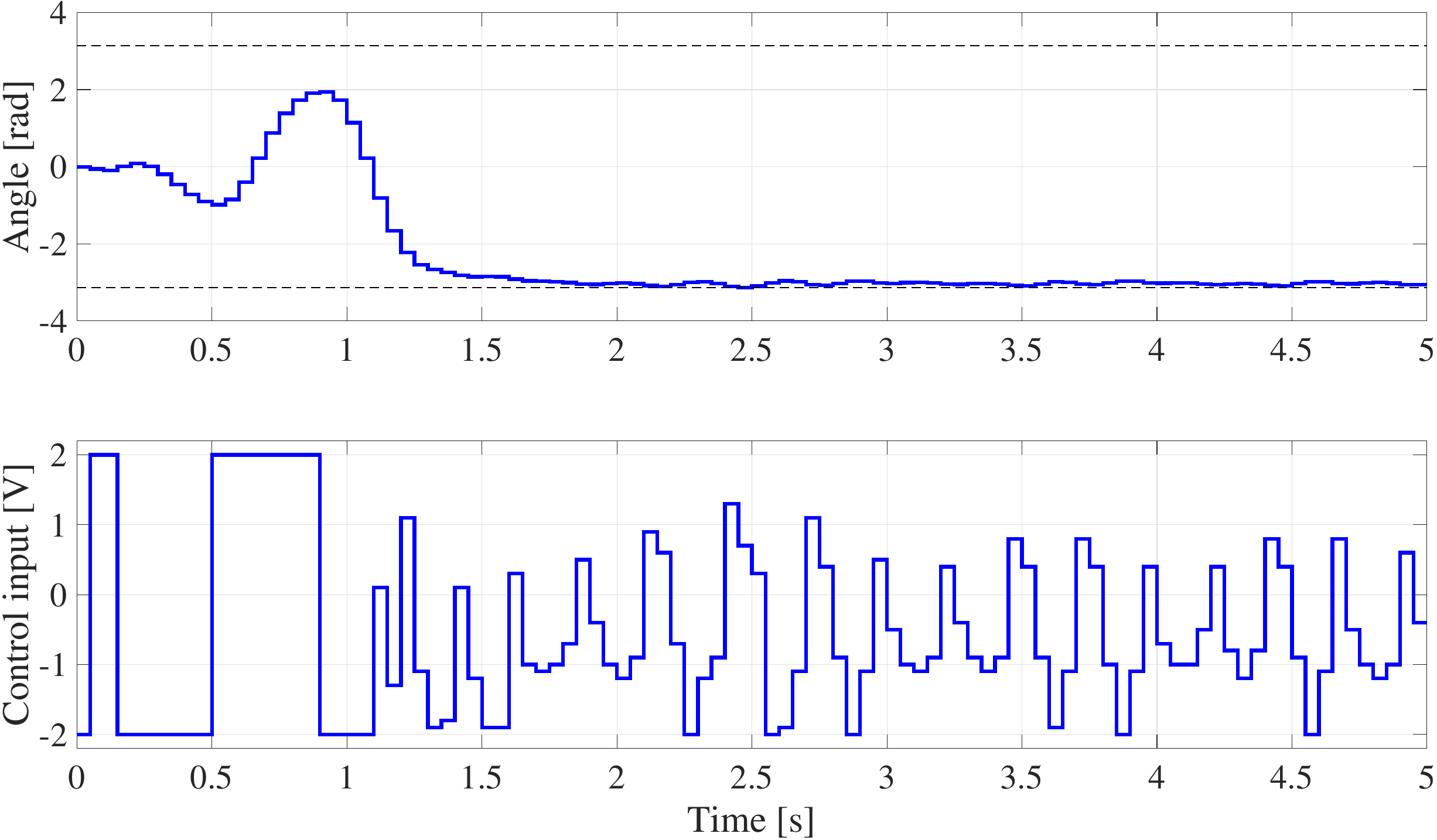}} \qquad
\subfigure[]{\includegraphics[width=0.47\columnwidth]{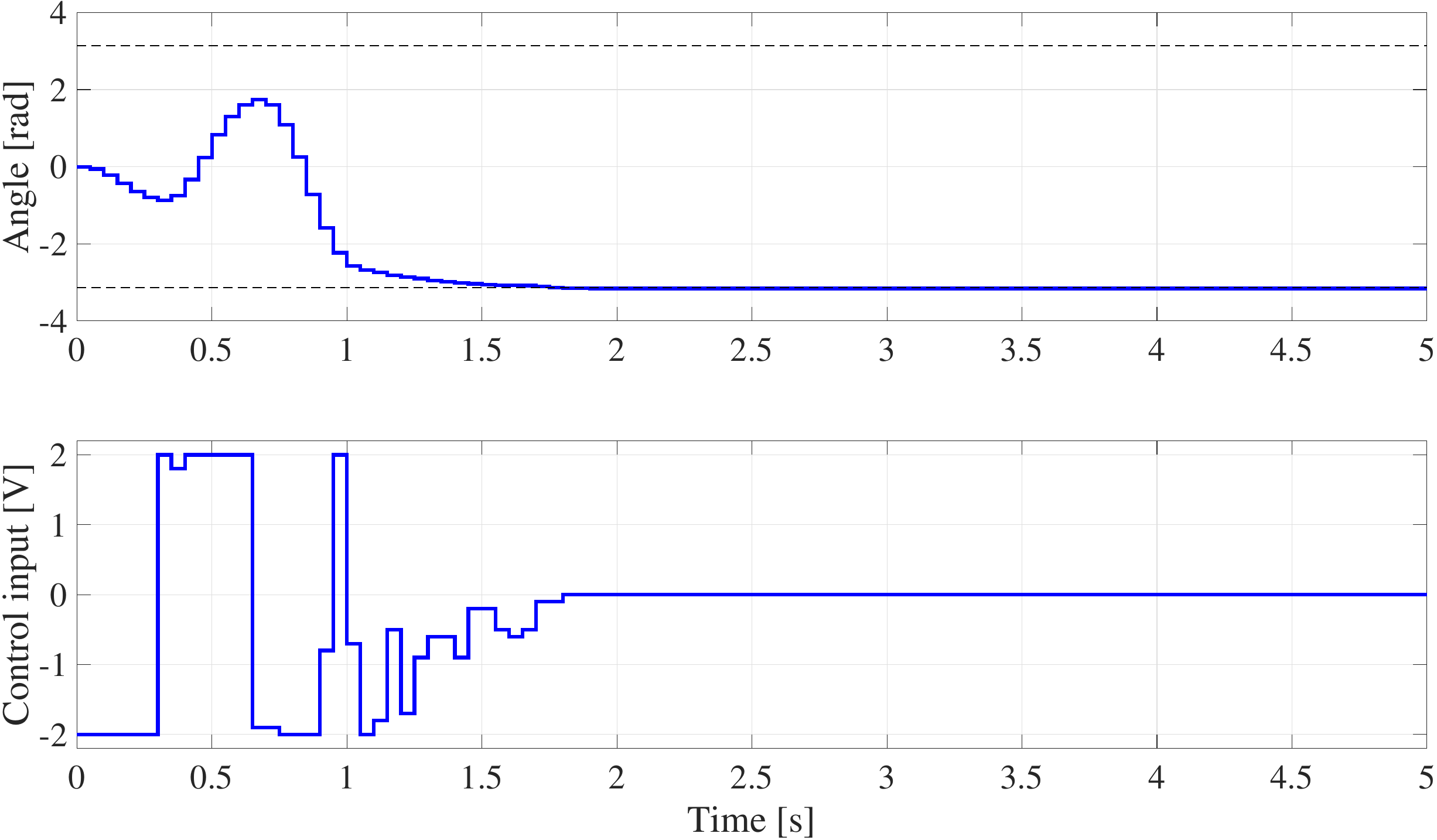}}
    \caption{A typical real swing-up experiment with the initial model (a) and the refined model (b).}
    \label{fig:rtswingup-initial-final}%
\end{figure}

\begin{figure}[htbp]
  \centering
  \includegraphics[width=0.6\columnwidth]{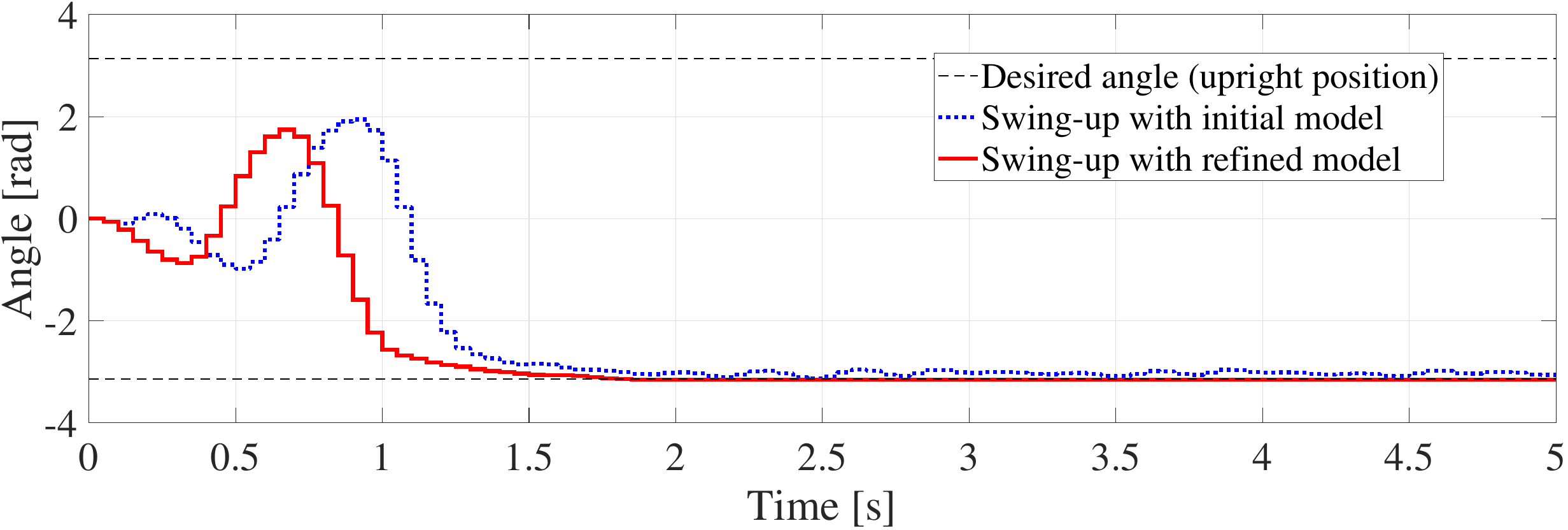}
  \caption{Comparison of the real swing-up response with the initial model, learnt from the random data, and the refined model, learnt from the random data merged with additional data from eight real swing-up experiments.}
  \label{fig:rtswingup-comparison}
\end{figure}

\begin{figure}[htbp]
  \centering
  \includegraphics[width=0.6\columnwidth]{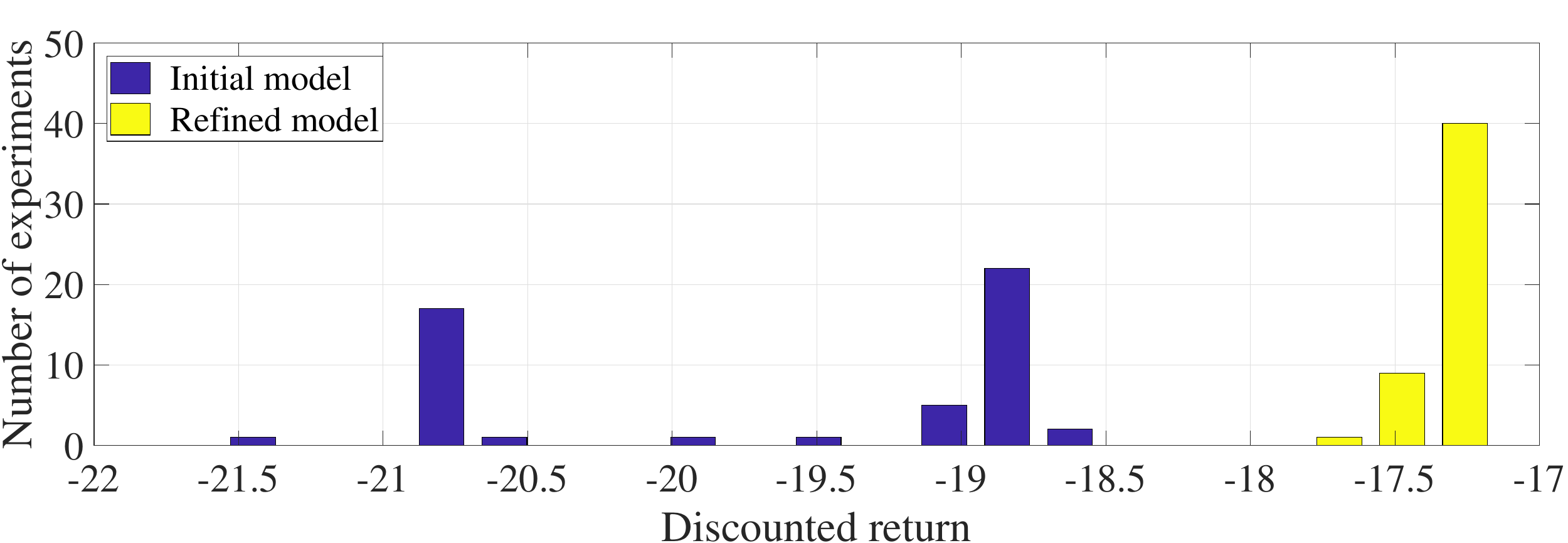}
  \caption{Histograms of 50 real experiments with the initial model and the refined model measured by the discounted return. The performance improvement is statistically significant ($p = 2 \times 10^{-22}$).}
  \label{fig:rtswingup-hist}
\end{figure}

In \emph{Experiment~C3}, we have shown that SR is able to find analytic process models using data collected on the real system. Already after a short (5~s) interaction under the random input, an analytic process model is found which enables RL to perform the swing-up, see Figure~\ref{fig:rtswingup-initial-final}a. Performing the swing-up task allows to collect more data in important parts of the state space around the trajectory to the goal state. Figure~\ref{fig:rtswingup-initial-final}b shows that the performance of the model further improves after adding data collected while performing the swing-up task with the initial model. Figure~\ref{fig:rtswingup-comparison} compares the swing-up response with the initial and the refined model. The histogram in Figure~\ref{fig:rtswingup-hist} and a two-sample t-test with unpooled variance applied to the discounted return show that the performance improvement between the policy based on the initial and the refined analytic process model is statistically significant ($p = 2 \times 10^{-22}$). The RMSE medians over 30 runs of the SNGP algorithm were $1.70 \times 10^{-2}$ for~$\alpha$ and $6.03 \times 10^{-1}$ for~$\dot\alpha$ in case of the initial model and $1.16 \times 10^{-2}$ for~$\alpha$ and $3.35 \times 10^{-1}$ for~$\dot\alpha$ in case of the refined model.

\subsubsection{Comparison with Alternative Methods}

We compared our modeling results with local linear regression (LLR) \cite{Atkeson1997}. We selected the Runge-Kutta data set with 1000 samples and zero noise as a reference training set and the regular grid as a test set (see Section \ref{sec:swing-up-data-sets} for details). The LLR memory contained 1000 samples and the number of nearest neighbors was set to~10. The RMSE achieved by LLR was $1.73 \times 10^{-1}$ for $\alpha$ and $6.93 \times 10^{0}$ for $\dot\alpha$. In both cases, the SNGP algorithm achieved a better RMSE by at least one order of magnitude ($6.11 \times 10^{-3}$ for $\alpha$ and $5.04 \times 10^{-1}$ for~$\dot\alpha$).

We also compared the results of our method to a neural network. Given the relative simplicity of the problem, the network had one hidden layer, consisting of 40 neurons, and it was trained using the Levenberg-Marquardt algorithm. The number of neurons in the hidden layer was tuned by testing networks with 5 to 100 neurons and choosing the one that performed best on the test data. The RMSE achieved on the aforementioned reference data set was $6.82 \times 10^{-2}$ for $\alpha$ and $2.59 \times 10^{0}$ for $\dot\alpha$. Again, compared to the RMSE values achieved by our method (stated at the end of the previous paragraph), symbolic regression finds substantially better models in terms of RMSE compared to those found by the neural network.

\section{Conclusions\label{sec:conclusions}}

We showed that symbolic regression is a very effective method for constructing dynamic process models from data. It generates parsimonious models in the form of analytic expressions, which makes it a good alternative to black-box models, especially in problems with limited amounts of data. Prior knowledge on the type of nonlinearities and model complexity can easily be included in the symbolic regression procedure. Despite the technique is not yet broadly used in the field of robotics and dynamic systems, we believe that it will become a standard tool for system identification.

The experiments with the walking robot demonstrate that symbolic regression can be used to construct precise process models even for high-dimensional systems. We have confirmed empirically that the computational complexity of the algorithm grows linearly with the dimensionality of the system. It is also worth mentioning that the complexity of the analytic models does not grow significantly with the complexity of the system.

The real-world experiment with the inverted pendulum shows that already after 5~seconds of interaction with the system, an initial analytic process model is found, which not only accurately predicts the process behavior, but also serves as a reliable model for the design of an RL controller. By collecting the data during several executions of the swing-up task using the initial analytic model and adding them to the data set used by SR to learn the model, the performance on the swing-up task further improves.

Our evaluation shows that two distinct symbolic regression algorithms, SNGP and MGGP, perform comparably well on the evaluated systems. This indicates that the proposed method is not dependent on the particular choice of the symbolic regression method. We compared the performance of symbolic regression with alternative state-of-the-art methods, in particular with neural networks and with local linear regression. The results show that the proposed method performs in most cases significantly better than the alternatives.

Another important outcome is that SR can be used to find both state-space and input--output models. The use of input--output models is beneficial because it does not require the observations of the full state vector and it also makes the algorithm faster because of modeling a reduced number of variables.

We have identified several possibilities for future extensions of this work. The main objective is to apply SR methods within the entire RL scheme, i.e., also for approximating the V-function, and also to use analytic models in combination with actor-critic online RL. In some cases, especially when using many features, analytic models tend to be unnecessarily complex. In our future work, we will investigate systematic reduction of analytic models.

\section*{Acknowledgments}

This work was supported by the European Regional Development Fund under the project Robotics for Industry 4.0 (reg.~no. CZ.02.1.01/0.0/0.0/15\_003/0000470) and by the Grant Agency of the Czech Technical University in Prague, grant no. SGS19/174/OHK3/3T/13.

The authors thank Tim de Bruin and Jon\'a\v{s} Kulh\'anek for their help with the DNN experiments for the walking robot and Jan \v{Z}egklitz for his help with experiments using the MGGP algorithm.

\newpage

\appendix

\section{Reinforcement Learning}
\label{sec:rl}

The system for which an optimal control strategy is to be learnt can be described by the nonlinear state-space model \eqref{eq:model} or the input--output (NARX) model \eqref{eq:model0-narx}. The following text describes the case using the state-space models. For the input--output models, the reward function, the return, the value function and the optimal control action \eqref{eq:reward}--\eqref{eq:hillclimb} can be defined analogously using the regressor $\varphi$ instead of the state variable $x$.

The  {\em reward function} assigns a scalar reward $r_{k+1} \in \mathbb{R}$ to the state transition from $x_k$ to $x_{k+1}$, under action $u_k$:
\begin{equation}
\begin{aligned}
r_{k+1} &= \rho(x_k, u_k, x_{k+1})\,.
\end{aligned}
\label{eq:reward}
\end{equation}
The reward function $\rho$ specifies the control goal, typically as the distance of the current state to a given goal state.

Based on model \eqref{eq:model}, we compute the optimal control policy $\pi: \mathcal{X} \rightarrow \mathcal{U}$ such that in each state it selects a control action so that the expected cumulative discounted reward over time, called the return, is maximized:
\begin{equation}
        R^{\pi} = E\Bigl\{\sum^{\infty}_{k=0}\gamma^{k}\rho\bigl(x_{k},\pi(x_{k}),x_{k+1}\bigr)\Bigr\}\,.
\label{eq:return}
\end{equation}
Here $\gamma\in (0,1)$ is a discount factor and the initial state $x_0$ is drawn from the state space domain $\mathcal{X}$ or its subset. Over the whole state space, the return is captured by the value function $V^{\pi}:\mathcal{X}\to \mathbb{R}$ defined as:
\begin{equation}
        V^\pi(x) = E\Bigl\{\sum^{\infty}_{k=0}\gamma^{k}\rho\bigl(x_{k},\pi(x_{k}),x_{k+1}\bigr)\Bigl\vert\,\Bigr. x_{0} = x\Bigr\}\,.
\label{eq:valuefunction}
\end{equation}
An approximation of the optimal V-function, denoted by $\hat{V}^*(x)$, can be computed by solving the Bellman optimality equation
\begin{equation}
\hat{V}^*(x) = \max_{u\in \mathcal{U}} \Bigl[\rho\bigl(x,u,f(x,u)\bigr)+\gamma \, \hat{V}^*\bigl(f (x,u)\bigr)\Bigr]\,.
\label{eq:BE}
\end{equation}
To simplify the notation, we drop the superscripts; $V(x)$ therefore denotes an approximation of the optimal V-function. Based on $V(x)$, the corresponding approximately optimal control action is found as the one that maximizes the right-hand side of (\ref{eq:BE}):
\begin{equation}
u = \argmax_{u'\in U} \left[\rho(x, u', f(x,u')) + \gamma \, V(f(x,u')) \right]\,.
\label{eq:hillclimb}
\end{equation}
In this work, the above equation is used online as the control policy $\pi$ with a set of discretized inputs $U$, so that the near-optimal control action can be found by enumeration.

\section{Detailed Experiment Results}
\label{sec:tables}

In this appendix, we present detailed results of the experiments described in Section~\ref{sec:experiments}.

\begin{table}[htpb]
\caption{Comparison of analytic process models for the experiment with the mobile robot. The table shows the RMSE medians over 30 runs of the SNGP (grey) and MGGP (white) algorithm for different numbers of features $n_f$ and different numbers of training samples $n_s$.}
\label{tab:mobile-robot}
\centering
\begin{footnotesize}
\begin{tabular}{|c|c|cccccc|}
\hline
\multirow{2}{*}{Variable} & \multirow{2}{*}{$n_f$} & \multicolumn{6}{c|}{Number of training samples $n_s$} \toppad \\
\hhline{~|~|-|-|-|-|-|-|}
 & & 20 & 50 & 100 & 200 & 500 & 1000 \toppad \\
\hline
\multirow{7}{*}{$x_{pos}$} & \multirow{2}{*}{1} & \gc{$1.77 \times 10^{-1}$} & \gc{$1.37 \times 10^{-1}$} & \gc{$9.98 \times 10^{-2}$} & \gc{$7.88 \times 10^{-1}$} & \gc{$3.25 \times 10^{-2}$} & \gcr{$2.87 \times 10^{-2}$} \toppad \\
 & & $9.03 \times 10^{-1}$ & $7.66 \times 10^{-1}$ & $6.97 \times 10^{-1}$ & $9.16 \times 10^{-1}$ & $3.88 \times 10^{-2}$ & $3.29 \times 10^{-2}$ \toppad \\
\hhline{~|-|-|-|-|-|-|-|}
 & \multirow{2}{*}{2} & \gc{$6.44 \times 10^{-8}$} & \gc{$3.19 \times 10^{-9}$} & \gc{$2.05 \times 10^{-9}$} & \gc{$3.55 \times 10^{-9}$} & \gc{$5.93 \times 10^{-9}$} & \gcr{$1.53 \times 10^{-9}$} \toppad \\
 & & $2.21 \times 10^{-9}$ & $5.64 \times 10^{-10}$ & $5.15 \times 10^{-6}$ & $1.11 \times 10^{-2}$ & $1.30 \times 10^{-10}$ & $1.37 \times 10^{-10}$ \toppad \\
\hhline{~|-|-|-|-|-|-|-|}
 & \multirow{2}{*}{10} & \gc{$3.78 \times 10^{-5}$} & \gc{$2.29 \times 10^{-7}$} & \gc{$1.09 \times 10^{-7}$} & \gc{$1.45 \times 10^{-7}$} & \gc{$5.21 \times 10^{-9}$} & \gcr{$2.68 \times 10^{-9}$} \toppad \\
 & & $2.37 \times 10^{-5}$ & $1.39 \times 10^{-7}$ & $8.50 \times 10^{-9}$ & $5.54 \times 10^{-9}$ & $2.70 \times 10^{-9}$ & $5.26 \times 10^{-10}$ \toppad \\
\hline
\multirow{7}{*}{$y_{pos}$} & \multirow{2}{*}{1} & \gc{$9.06 \times 10^{-1}$} & \gc{$4.34 \times 10^{-1}$} & \gc{$1.39 \times 10^{-1}$} & \gc{$1.74 \times 10^{-1}$} & \gc{$3.21 \times 10^{-2}$} & \gcr{$3.16 \times 10^{-2}$} \toppad \\
 & & $8.74 \times 10^{-1}$ & $9.47 \times 10^{-1}$ & $7.75 \times 10^{-1}$ & $7.33 \times 10^{-1}$ & $3.31 \times 10^{-2}$ & $3.16 \times 10^{-2}$ \toppad \\
\hhline{~|-|-|-|-|-|-|-|}
 & \multirow{2}{*}{2} & \gc{$4.87 \times 10^{-1}$} & \gc{$1.81 \times 10^{-8}$} & \gc{$1.18 \times 10^{-8}$} & \gc{$4.09 \times 10^{-9}$} & \gc{$1.93 \times 10^{-8}$} & \gcr{$2.39 \times 10^{-8}$} \toppad \\
 & & $3.39 \times 10^{-1}$ & $3.38 \times 10^{-2}$ & $2.89 \times 10^{-10}$ & $2.76 \times 10^{-10}$ & $2.68 \times 10^{-10}$ & $2.14 \times 10^{-2}$ \toppad \\
\hhline{~|-|-|-|-|-|-|-|}
 & \multirow{2}{*}{10} & \gc{$4.48 \times 10^{-4}$} & \gc{$2.04 \times 10^{-7}$} & \gc{$4.11 \times 10^{-7}$} & \gc{$1.91 \times 10^{-7}$} & \gc{$1.60 \times 10^{-8}$} & \gcr{$1.25 \times 10^{-8}$} \toppad \\
 & & $9.32 \times 10^{-5}$ & $1.16 \times 10^{-7}$ & $7.54 \times 10^{-9}$ & $6.45 \times 10^{-9}$ & $2.34 \times 10^{-9}$ & $8.33 \times 10^{-10}$ \toppad \\
\hline
\multirow{7}{*}{$\phi$} & \multirow{2}{*}{1} & \gc{$9.81 \times 10^{-2}$} & \gc{$2.60 \times 10^{-2}$} & \gc{$6.44 \times 10^{-4}$} & \gc{$6.57 \times 10^{-5}$} & \gc{$6.79 \times 10^{-4}$} & \gcr{$5.55 \times 10^{-3}$} \toppad \\
 & & $3.38 \times 10^{0}$ & $3.19 \times 10^{0}$ & $2.49 \times 10^{-2}$ & $1.34 \times 10^{-3}$ & $5.51 \times 10^{-5}$ & $5.48 \times 10^{-5}$ \toppad \\
\hhline{~|-|-|-|-|-|-|-|}
 & \multirow{2}{*}{2} & \gc{$7.05 \times 10^{-8}$} & \gc{$1.36 \times 10^{-8}$} & \gc{$6.16 \times 10^{-9}$} & \gc{$3.78 \times 10^{-8}$} & \gc{$7.78 \times 10^{-9}$} & \gcr{$5.16 \times 10^{-8}$} \toppad \\
 & & $1.47 \times 10^{-9}$ & $5.08 \times 10^{-10}$ & $4.16 \times 10^{-10}$ & $4.02 \times 10^{-10}$ & $4.00 \times 10^{-10}$ & $4.01 \times 10^{-10}$ \toppad \\
\hhline{~|-|-|-|-|-|-|-|}
 & \multirow{2}{*}{10} & \gc{$5.35 \times 10^{-6}$} & \gc{$1.85 \times 10^{-6}$} & \gc{$2.09 \times 10^{-6}$} & \gc{$4.01 \times 10^{-8}$} & \gc{$6.00 \times 10^{-9}$} & \gcr{$3.69 \times 10^{-8}$} \toppad \\
 & & $6.45 \times 10^{-8}$ & $9.34 \times 10^{-9}$ & $2.87 \times 10^{-9}$ & $1.35 \times 10^{-9}$ & $4.07 \times 10^{-10}$ & $4.00 \times 10^{-10}$ \toppad \\
\hline
\end{tabular}
\end{footnotesize}
\end{table}

\begin{table}[htbp]
\caption{Comparison of the state-space analytic process models for the walking robot LEO in \emph{Experiment~B1}. The table shows the RMSE medians over 30 runs of the SNGP algorithm for varying number of features $n_f$ and number of training samples $n_s$.}
\centering
\label{tab:walking-robot-stsp}
\begin{footnotesize}
\begin{tabular}{|c|c|cccccc|}
\hline
\multirow{2}{*}{Variable} & \multirow{2}{*}{$n_f$} & \multicolumn{6}{c|}{Number of training samples $n_s$}  \toppad \\
\cline{3-8}
 & & 100 & 200 & 500 & 1000 & 2000 & 5000 \toppad \\
\hline
\multirow{3}{*}{$\psi_{TRS}$} & 1 & $4.09 \times 10^{-2}$ & $1.72 \times 10^{-2}$ & $4.15 \times 10^{-2}$ & $5.38 \times 10^{-2}$ & $5.46 \times 10^{-2}$ & $5.68 \times 10^{-2}$ \toppad \\
\cline{2-8}
 & 5 & $1.90 \times 10^{-2}$ & $1.55 \times 10^{-2}$ & $1.46 \times 10^{-2}$ & $1.40 \times 10^{-2}$ & $1.39 \times 10^{-2}$ & $1.38 \times 10^{-2}$ \toppad \\
\cline{2-8}
 & 10 & $2.01 \times 10^{-2}$ & $1.62 \times 10^{-2}$ & $1.44 \times 10^{-2}$ & $1.32 \times 10^{-2}$ & $1.29 \times 10^{-2}$ & $1.25 \times 10^{-2}$ \toppad \\
\hline
\multirow{3}{*}{$\psi_{LH}$} & 1 & $2.99 \times 10^{-2}$ & $2.99 \times 10^{-2}$ & $3.60 \times 10^{-2}$ & $2.24 \times 10^{-2}$ & $2.34 \times 10^{-2}$ & $8.81 \times 10^{-2}$ \toppad \\
\cline{2-8}
 & 5 & $2.78 \times 10^{-2}$ & $2.38 \times 10^{-2}$ & $2.15 \times 10^{-2}$ & $2.07 \times 10^{-2}$ & $2.02 \times 10^{-2}$ & $2.01 \times 10^{-2}$ \toppad \\
\cline{2-8}
 & 10 & $2.99 \times 10^{-2}$ & $2.53 \times 10^{-2}$ & $2.20 \times 10^{-2}$ & $2.06 \times 10^{-2}$ & $1.95 \times 10^{-2}$ & $1.92 \times 10^{-2}$ \toppad \\
\hline
\multirow{3}{*}{$\psi_{RH}$} & 1 & $1.05 \times 10^{-1}$ & $9.16 \times 10^{-2}$ & $4.01 \times 10^{-2}$ & $3.71 \times 10^{-2}$ & $3.36 \times 10^{-2}$ & $2.84 \times 10^{-2}$ \toppad \\
\cline{2-8}
 & 5 & $3.81 \times 10^{-2}$ & $3.19 \times 10^{-2}$ & $2.69 \times 10^{-2}$ & $2.71 \times 10^{-2}$ & $2.61 \times 10^{-2}$ & $2.56 \times 10^{-2}$ \toppad \\
\cline{2-8}
 & 10 & $4.15 \times 10^{-2}$ & $3.54 \times 10^{-2}$ & $2.65 \times 10^{-2}$ & $2.65 \times 10^{-2}$ & $2.46 \times 10^{-2}$ & $2.46 \times 10^{-2}$ \toppad \\
\hline
\multirow{3}{*}{$\psi_{LK}$} & 1 & $5.52 \times 10^{-2}$ & $8.01 \times 10^{-2}$ & $2.43 \times 10^{-2}$ & $2.35 \times 10^{-2}$ & $2.40 \times 10^{-2}$ & $2.28 \times 10^{-2}$ \toppad \\
\cline{2-8}
 & 5 & $3.10 \times 10^{-2}$ & $2.68 \times 10^{-2}$ & $2.29 \times 10^{-2}$ & $2.15 \times 10^{-2}$ & $2.06 \times 10^{-2}$ & $2.07 \times 10^{-2}$ \toppad \\
\cline{2-8}
 & 10 & $3.43 \times 10^{-2}$ & $2.87 \times 10^{-2}$ & $2.22 \times 10^{-2}$ & $2.10 \times 10^{-2}$ & $1.97 \times 10^{-2}$ & $1.89 \times 10^{-2}$ \toppad \\
\hline
\multirow{3}{*}{$\psi_{RK}$} & 1 & $2.82 \times 10^{-2}$ & $2.36 \times 10^{-2}$ & $2.31 \times 10^{-2}$ & $2.31 \times 10^{-2}$ & $2.15 \times 10^{-2}$ & $2.13 \times 10^{-2}$ \toppad \\
\cline{2-8}
 & 5 & $2.77 \times 10^{-2}$ & $2.28 \times 10^{-2}$ & $2.01 \times 10^{-2}$ & $2.01 \times 10^{-2}$ & $2.01 \times 10^{-2}$ & $1.90 \times 10^{-2}$ \toppad \\
\cline{2-8}
 & 10 & $3.08 \times 10^{-2}$ & $2.45 \times 10^{-2}$ & $1.96 \times 10^{-2}$ & $1.87 \times 10^{-2}$ & $1.86 \times 10^{-2}$ & $1.77 \times 10^{-2}$ \toppad \\
\hline
\multirow{3}{*}{$\psi_{LA}$} & 1 & $9.31 \times 10^{-2}$ & $1.11 \times 10^{-1}$ & $1.10 \times 10^{-1}$ & $1.10 \times 10^{-1}$ & $7.07 \times 10^{-2}$ & $5.74 \times 10^{-2}$ \toppad \\
\cline{2-8}
 & 5 & $5.18 \times 10^{-2}$ & $4.00 \times 10^{-2}$ & $3.11 \times 10^{-2}$ & $2.95 \times 10^{-2}$ & $2.80 \times 10^{-2}$ & $2.81 \times 10^{-2}$ \toppad \\
\cline{2-8}
 & 10 & $5.66 \times 10^{-2}$ & $4.31 \times 10^{-2}$ & $3.16 \times 10^{-2}$ & $2.92 \times 10^{-2}$ & $2.73 \times 10^{-2}$ & $2.62 \times 10^{-2}$ \toppad \\
\hline
\multirow{3}{*}{$\psi_{RA}$} & 1 & $4.65 \times 10^{-2}$ & $4.51 \times 10^{-2}$ & $4.24 \times 10^{-2}$ & $4.54 \times 10^{-2}$ & $4.33 \times 10^{-2}$ & $7.37 \times 10^{-2}$ \toppad \\
\cline{2-8}
 & 5 & $4.98 \times 10^{-2}$ & $4.49 \times 10^{-2}$ & $3.77 \times 10^{-2}$ & $3.66 \times 10^{-2}$ & $3.65 \times 10^{-2}$ & $3.52 \times 10^{-2}$ \toppad \\
\cline{2-8}
 & 10 & $5.35 \times 10^{-2}$ & $4.70 \times 10^{-2}$ & $3.84 \times 10^{-2}$ & $3.65 \times 10^{-2}$ & $3.50 \times 10^{-2}$ & $3.39 \times 10^{-2}$ \toppad \\
\hline
\multirow{3}{*}{$\dot\psi_{TRS}$} & 1 & $8.91 \times 10^{-1}$ & $8.51 \times 10^{-1}$ & $8.19 \times 10^{-1}$ & $7.99 \times 10^{-1}$ & $7.94 \times 10^{-1}$ & $7.84 \times 10^{-1}$ \toppad \\
\cline{2-8}
 & 5 & $1.07 \times 10^{0}$ & $8.72 \times 10^{-1}$ & $7.78 \times 10^{-1}$ & $7.19 \times 10^{-1}$ & $6.92 \times 10^{-1}$ & $6.86 \times 10^{-1}$ \toppad \\
\cline{2-8}
 & 10 & $1.20 \times 10^{0}$ & $9.27 \times 10^{-1}$ & $7.93 \times 10^{-1}$ & $7.00 \times 10^{-1}$ & $6.67 \times 10^{-1}$ & $6.41 \times 10^{-1}$ \toppad \\
\hline
\multirow{3}{*}{$\dot\psi_{LH}$} & 1 & $1.44 \times 10^{0}$ & $1.23 \times 10^{0}$ & $1.16 \times 10^{0}$ & $1.15 \times 10^{0}$ & $1.14 \times 10^{0}$ & $1.14 \times 10^{0}$ \toppad \\
\cline{2-8}
 & 5 & $2.22 \times 10^{0}$ & $1.41 \times 10^{0}$ & $1.17 \times 10^{0}$ & $1.15 \times 10^{0}$ & $1.11 \times 10^{0}$ & $1.08 \times 10^{0}$ \toppad \\
\cline{2-8}
 & 10 & $2.07 \times 10^{0}$ & $1.48 \times 10^{0}$ & $1.20 \times 10^{0}$ & $1.16 \times 10^{0}$ & $1.10 \times 10^{0}$ & $1.06 \times 10^{0}$ \toppad \\
\hline
\multirow{3}{*}{$\dot\psi_{RH}$} & 1 & $1.49 \times 10^{0}$ & $1.32 \times 10^{0}$ & $1.31 \times 10^{0}$ & $1.28 \times 10^{0}$ & $1.25 \times 10^{0}$ & $1.24 \times 10^{0}$ \toppad \\
\cline{2-8}
 & 5 & $1.92 \times 10^{0}$ & $1.47 \times 10^{0}$ & $1.38 \times 10^{0}$ & $1.25 \times 10^{0}$ & $1.17 \times 10^{0}$ & $1.14 \times 10^{0}$ \toppad \\
\cline{2-8}
 & 10 & $1.97 \times 10^{0}$ & $1.57 \times 10^{0}$ & $1.52 \times 10^{0}$ & $1.27 \times 10^{0}$ & $1.17 \times 10^{0}$ & $1.12 \times 10^{0}$ \toppad \\
\hline
\multirow{3}{*}{$\dot\psi_{LK}$} & 1 & $1.59 \times 10^{0}$ & $1.25 \times 10^{0}$ & $1.14 \times 10^{0}$ & $1.11 \times 10^{0}$ & $1.10 \times 10^{0}$ & $1.09 \times 10^{0}$ \toppad \\
\cline{2-8}
 & 5 & $1.79 \times 10^{0}$ & $1.47 \times 10^{0}$ & $1.15 \times 10^{0}$ & $1.09 \times 10^{0}$ & $1.05 \times 10^{0}$ & $9.94 \times 10^{-1}$ \toppad \\
\cline{2-8}
 & 10 & $1.90 \times 10^{0}$ & $1.57 \times 10^{0}$ & $1.20 \times 10^{0}$ & $1.11 \times 10^{0}$ & $1.08 \times 10^{0}$ & $9.87 \times 10^{-1}$ \toppad \\
\hline
\multirow{3}{*}{$\dot\psi_{RK}$} & 1 & $1.02 \times 10^{0}$ & $9.35 \times 10^{-1}$ & $9.18 \times 10^{-1}$ & $9.24 \times 10^{-1}$ & $9.16 \times 10^{-1}$ & $9.05 \times 10^{-1}$ \toppad \\
\cline{2-8}
 & 5 & $1.13 \times 10^{0}$ & $9.98 \times 10^{-1}$ & $9.40 \times 10^{-1}$ & $9.29 \times 10^{-1}$ & $8.64 \times 10^{-1}$ & $8.32 \times 10^{-1}$ \toppad \\
\cline{2-8}
 & 10 & $1.24 \times 10^{0}$ & $1.07 \times 10^{0}$ & $9.83 \times 10^{-1}$ & $9.63 \times 10^{-1}$ & $8.73 \times 10^{-1}$ & $8.20 \times 10^{-1}$ \toppad \\
\hline
\multirow{3}{*}{$\dot\psi_{LA}$} & 1 & $1.76 \times 10^{0}$ & $1.52 \times 10^{0}$ & $1.32 \times 10^{0}$ & $1.31 \times 10^{0}$ & $1.28 \times 10^{0}$ & $1.26 \times 10^{0}$ \toppad \\
\cline{2-8}
 & 5 & $2.01 \times 10^{0}$ & $1.63 \times 10^{0}$ & $1.29 \times 10^{0}$ & $1.24 \times 10^{0}$ & $1.14 \times 10^{0}$ & $1.10 \times 10^{0}$ \toppad \\
\cline{2-8}
 & 10 & $2.18 \times 10^{0}$ & $1.67 \times 10^{0}$ & $1.35 \times 10^{0}$ & $1.25 \times 10^{0}$ & $1.15 \times 10^{0}$ & $1.10 \times 10^{0}$ \toppad \\
\hline
\multirow{3}{*}{$\dot\psi_{RA}$} & 1 & $1.69 \times 10^{0}$ & $1.64 \times 10^{0}$ & $1.60 \times 10^{0}$ & $1.58 \times 10^{0}$ & $1.58 \times 10^{0}$ & $1.58 \times 10^{0}$ \toppad \\
\cline{2-8}
 & 5 & $1.84 \times 10^{0}$ & $1.75 \times 10^{0}$ & $1.52 \times 10^{0}$ & $1.48 \times 10^{0}$ & $1.43 \times 10^{0}$ & $1.38 \times 10^{0}$ \toppad \\
\cline{2-8}
 & 10 & $1.92 \times 10^{0}$ & $1.86 \times 10^{0}$ & $1.62 \times 10^{0}$ & $1.51 \times 10^{0}$ & $1.43 \times 10^{0}$ & $1.35 \times 10^{0}$ \toppad \\
\hline
\end{tabular}
\end{footnotesize}
\end{table}

\begin{table}[htbp]
\caption{Comparison of the input--output analytic process models for the walking robot LEO in \emph{Experiment~B2}. The table shows the RMSE medians over 30 runs of the SNGP algorithm for varying number of features $n_f$ and number of training samples $n_s$.}
\label{tab:walking-robot-narx}
\centering
\begin{footnotesize}
\begin{tabular}{|c|c|cccccc|}
\hline
\multirow{2}{*}{Variable} & \multirow{2}{*}{$n_f$} & \multicolumn{6}{c|}{Number of training samples $n_s$}  \toppad \\
\cline{3-8}
 & & 100 & 200 & 500 & 1000 & 2000 & 5000 \toppad \\
\hline
\multirow{3}{*}{$\psi_{TRS}$} & 1 & $5.78 \times 10^{-2}$ & $5.77 \times 10^{-2}$ & $5.71 \times 10^{-2}$ & $5.58 \times 10^{-2}$ & $5.60 \times 10^{-2}$ & $5.69 \times 10^{-2}$ \toppad \\
\cline{2-8}
 & 5 & $3.84 \times 10^{-2}$ & $3.60 \times 10^{-2}$ & $3.15 \times 10^{-2}$ & $2.65 \times 10^{-2}$ & $2.45 \times 10^{-2}$ & $2.33 \times 10^{-2}$ \toppad \\
\cline{2-8}
 & 10 & $4.07 \times 10^{-2}$ & $3.36 \times 10^{-2}$ & $2.88 \times 10^{-2}$ & $2.48 \times 10^{-2}$ & $2.09 \times 10^{-2}$ & $2.06 \times 10^{-2}$ \toppad \\
\hline
\multirow{3}{*}{$\psi_{LH}$} & 1 & $6.75 \times 10^{-2}$ & $6.57 \times 10^{-2}$ & $6.57 \times 10^{-2}$ & $5.90 \times 10^{-2}$ & $1.07 \times 10^{-1}$ & $6.67 \times 10^{-2}$ \toppad \\
\cline{2-8}
 & 5 & $3.80 \times 10^{-2}$ & $2.97 \times 10^{-2}$ & $2.62 \times 10^{-2}$ & $2.71 \times 10^{-2}$ & $2.56 \times 10^{-2}$ & $2.53 \times 10^{-2}$ \toppad \\
\cline{2-8}
 & 10 & $3.85 \times 10^{-2}$ & $3.15 \times 10^{-2}$ & $2.65 \times 10^{-2}$ & $2.64 \times 10^{-2}$ & $2.46 \times 10^{-2}$ & $2.40 \times 10^{-2}$ \toppad \\
\hline
\multirow{3}{*}{$\psi_{RH}$} & 1 & $8.04 \times 10^{-2}$ & $8.57 \times 10^{-2}$ & $6.62 \times 10^{-2}$ & $1.14 \times 10^{-1}$ & $1.06 \times 10^{-1}$ & $7.03 \times 10^{-2}$ \toppad \\
\cline{2-8}
 & 5 & $4.92 \times 10^{-2}$ & $3.81 \times 10^{-2}$ & $3.29 \times 10^{-2}$ & $3.20 \times 10^{-2}$ & $3.11 \times 10^{-2}$ & $3.04 \times 10^{-2}$ \toppad \\
\cline{2-8}
 & 10 & $5.40 \times 10^{-2}$ & $4.01 \times 10^{-2}$ & $3.25 \times 10^{-2}$ & $3.08 \times 10^{-2}$ & $2.88 \times 10^{-2}$ & $2.85 \times 10^{-2}$ \toppad \\
\hline
\multirow{3}{*}{$\psi_{LK}$} & 1 & $8.06 \times 10^{-2}$ & $5.52 \times 10^{-2}$ & $8.22 \times 10^{-2}$ & $7.52 \times 10^{-2}$ & $7.52 \times 10^{-2}$ & $5.77 \times 10^{-2}$ \toppad \\
\cline{2-8}
 & 5 & $3.66 \times 10^{-2}$ & $2.95 \times 10^{-2}$ & $2.46 \times 10^{-2}$ & $2.31 \times 10^{-2}$ & $2.20 \times 10^{-2}$ & $2.10 \times 10^{-2}$ \toppad \\
\cline{2-8}
 & 10 & $3.96 \times 10^{-2}$ & $3.09 \times 10^{-2}$ & $2.44 \times 10^{-2}$ & $2.21 \times 10^{-2}$ & $2.09 \times 10^{-2}$ & $2.04 \times 10^{-2}$ \toppad \\
\hline
\multirow{3}{*}{$\psi_{RK}$} & 1 & $8.69 \times 10^{-2}$ & $3.65 \times 10^{-2}$ & $2.99 \times 10^{-2}$ & $2.56 \times 10^{-2}$ & $8.88 \times 10^{-2}$ & $3.83 \times 10^{-2}$ \toppad \\
\cline{2-8}
 & 5 & $3.20 \times 10^{-2}$ & $2.62 \times 10^{-2}$ & $2.36 \times 10^{-2}$ & $2.26 \times 10^{-2}$ & $2.20 \times 10^{-2}$ & $2.18 \times 10^{-2}$ \toppad \\
\cline{2-8}
 & 10 & $3.49 \times 10^{-2}$ & $2.76 \times 10^{-2}$ & $2.24 \times 10^{-2}$ & $2.16 \times 10^{-2}$ & $2.08 \times 10^{-2}$ & $2.01 \times 10^{-2}$ \toppad \\
\hline
\multirow{3}{*}{$\psi_{LA}$} & 1 & $7.50 \times 10^{-2}$ & $1.07 \times 10^{-1}$ & $4.41 \times 10^{-2}$ & $1.03 \times 10^{-1}$ & $5.85 \times 10^{-2}$ & $1.03 \times 10^{-1}$ \toppad \\
\cline{2-8}
 & 5 & $5.44 \times 10^{-2}$ & $4.22 \times 10^{-2}$ & $3.27 \times 10^{-2}$ & $3.00 \times 10^{-2}$ & $2.89 \times 10^{-2}$ & $2.79 \times 10^{-2}$ \toppad \\
\cline{2-8}
 & 10 & $5.94 \times 10^{-2}$ & $4.62 \times 10^{-2}$ & $3.26 \times 10^{-2}$ & $2.96 \times 10^{-2}$ & $2.75 \times 10^{-2}$ & $2.66 \times 10^{-2}$ \toppad \\
\hline
\multirow{3}{*}{$\psi_{RA}$} & 1 & $1.19 \times 10^{-1}$ & $1.20 \times 10^{-1}$ & $9.65 \times 10^{-2}$ & $9.63 \times 10^{-2}$ & $1.02 \times 10^{-1}$ & $5.11 \times 10^{-2}$ \toppad \\
\cline{2-8}
 & 5 & $5.10 \times 10^{-2}$ & $4.45 \times 10^{-2}$ & $3.79 \times 10^{-2}$ & $3.69 \times 10^{-2}$ & $3.65 \times 10^{-2}$ & $3.60 \times 10^{-2}$ \toppad \\
\cline{2-8}
 & 10 & $5.51 \times 10^{-2}$ & $4.49 \times 10^{-2}$ & $3.79 \times 10^{-2}$ & $3.59 \times 10^{-2}$ & $3.44 \times 10^{-2}$ & $3.38 \times 10^{-2}$ \toppad \\
\hline
\end{tabular}
\end{footnotesize}
\end{table}

\begin{table}[htbp]
\caption{Comparison of the RMSE of the state-space process models calculated on the test data set for the walking robot LEO using two variants of a deep neural network (DNN-A and DNN-B) and SNGP. The reference configuration of SNGP used for this comparison was $n_f = 10$ and $n_s = 1000$.}
\label{tab:walking-robot-dnn}
\centering
\begin{footnotesize}
\begin{tabular}{|c|ccc|}
\hline
\multirow{2}{*}{Variable} & \multicolumn{3}{c|}{Method} \toppad \\
\cline{2-4}
 & DNN-A & DNN-B & SNGP \toppad \\
\hline
$\psi_{TRS}$ & $1.33 \times 10^{-1}$ & $9.27 \times 10^{-2}$ & $1.32 \times 10^{-2}$ \toppad \\
\hline
$\psi_{LH}$ & $1.86 \times 10^{-1}$ & $1.54 \times 10^{-1}$ & $2.06 \times 10^{-2}$ \toppad \\
\hline
$\psi_{RH}$ & $2.08 \times 10^{-1}$ & $1.23 \times 10^{-1}$ & $2.65 \times 10^{-2}$ \toppad \\
\hline
$\psi_{LK}$ & $2.24 \times 10^{-1}$ & $1.37 \times 10^{-1}$ & $2.10 \times 10^{-2}$ \toppad \\
\hline
$\psi_{RK}$ & $2.02 \times 10^{-1}$ & $1.10 \times 10^{-1}$ & $1.87 \times 10^{-2}$ \toppad \\
\hline
$\psi_{LA}$ & $1.62 \times 10^{-1}$ & $1.24 \times 10^{-1}$ & $2.92 \times 10^{-2}$ \toppad \\
\hline
$\psi_{RA}$ & $1.54 \times 10^{-1}$ & $9.36 \times 10^{-2}$ & $3.65 \times 10^{-2}$ \toppad \\
\hline
$\dot\psi_{TRS}$ & $7.39 \times 10^{-1}$ & $6.38 \times 10^{-1}$ & $7.00 \times 10^{-1}$ \toppad \\
\hline
$\dot\psi_{LH}$ & $1.13 \times 10^{0}$ & $1.12 \times 10^{0}$ & $1.16 \times 10^{0}$ \toppad \\
\hline
$\dot\psi_{RH}$ & $1.22 \times 10^{0}$ & $1.20 \times 10^{0}$ & $1.27 \times 10^{0}$ \toppad \\
\hline
$\dot\psi_{LK}$ & $1.08 \times 10^{0}$ & $1.06 \times 10^{0}$ & $1.11 \times 10^{0}$ \toppad \\
\hline
$\dot\psi_{RK}$ & $9.49 \times 10^{-1}$ & $8.68 \times 10^{-1}$ & $9.63 \times 10^{-1}$ \toppad \\
\hline
$\dot\psi_{LA}$ & $1.23 \times 10^{0}$ & $1.23 \times 10^{0}$ & $1.25 \times 10^{0}$ \toppad \\
\hline
$\dot\psi_{RA}$ & $1.54 \times 10^{0}$ & $1.42 \times 10^{0}$ & $1.51 \times 10^{0}$ \toppad \\
\hline
\end{tabular}
\end{footnotesize}
\end{table}

\begin{table}
\caption{Comparison of analytic process models for the inverted pendulum system in \emph{Experiment~C1}. The table shows the RMSE medians over 30 runs of the SNGP (grey) and MGGP (white) algorithm for varying number of features $n_f$ and varying number of training samples $n_s$.
\label{tab:e2}} \centering
\begin{footnotesize}
\begin{tabular}{|c|c|cccccc|}
\hline
\multirow{2}{*}{Variable} & \multirow{2}{*}{$n_f$} & \multicolumn{6}{c|}{Number of training samples $n_s$} \toppad \\
\hhline{~|~|-|-|-|-|-|-|}
 & & 20 & 50 & 100 & 200 & 500 & 1000 \toppad \\
\hline
\multirow{7}{*}{$\alpha$} & \multirow{2}{*}{1} & \gc{$9.19 \times 10^{-4}$} & \gc{$3.80 \times 10^{-4}$} & \gc{$2.45 \times 10^{-2}$} & \gc{$2.28 \times 10^{-3}$} & \gc{$1.89 \times 10^{-3}$} & \gcr{$2.38 \times 10^{-3}$} \toppad \\
 & & $5.51 \times 10^{-1}$ & $3.33 \times 10^{-1}$ & $4.46 \times 10^{-1}$ & $3.15 \times 10^{-1}$ & $2.44 \times 10^{-1}$ & $3.25 \times 10^{-1}$ \toppad \\
\hhline{~|-|-|-|-|-|-|-|}
 & \multirow{2}{*}{2} & \gc{$2.09 \times 10^{-7}$} & \gc{$2.39 \times 10^{-7}$} & \gc{$1.06 \times 10^{-7}$} & \gc{$1.82 \times 10^{-9}$} & \gc{$1.94 \times 10^{-8}$} & \gcr{$4.60 \times 10^{-9}$} \toppad \\
 & & $3.94 \times 10^{-10}$ & $3.78 \times 10^{-10}$ & $3.77 \times 10^{-10}$ & $3.76 \times 10^{-10}$ & $3.76 \times 10^{-10}$ & $3.76 \times 10^{-10}$ \toppad \\
\hhline{~|-|-|-|-|-|-|-|}
 & \multirow{2}{*}{10} & \gc{$5.03 \times 10^{-9}$} & \gc{$4.44 \times 10^{-7}$} & \gc{$4.41 \times 10^{-9}$} & \gc{$1.35 \times 10^{-9}$} & \gc{$8.45 \times 10^{-10}$} & \gcr{$4.56 \times 10^{-10}$} \toppad \\
 & & $4.29 \times 10^{-10}$ & $3.87 \times 10^{-10}$ & $3.87 \times 10^{-10}$ & $3.80 \times 10^{-10}$ & $3.77 \times 10^{-10}$ & $3.76 \times 10^{-10}$ \toppad \\
\hline
\multirow{7}{*}{$\dot\alpha$} & \multirow{2}{*}{1} & \gc{$7.97 \times 10^{-1}$} & \gc{$3.17 \times 10^{-1}$} & \gc{$2.51 \times 10^{-1}$} & \gc{$2.61 \times 10^{-1}$} & \gc{$2.34 \times 10^{-1}$} & \gcr{$5.11 \times 10^{-1}$} \toppad \\
 & & $9.21 \times 10^{-1}$ & $3.64 \times 10^{-1}$ & $2.42 \times 10^{-1}$ & $1.52 \times 10^{-1}$ & $3.42 \times 10^{-1}$ & $2.14 \times 10^{-1}$ \toppad \\
\hhline{~|-|-|-|-|-|-|-|}
 & \multirow{2}{*}{4} & \gc{$1.12 \times 10^{-6}$} & \gc{$5.61 \times 10^{-7}$} & \gc{$1.19 \times 10^{-6}$} & \gc{$1.61 \times 10^{-6}$} & \gc{$8.17 \times 10^{-7}$} & \gcr{$6.75 \times 10^{-7}$} \toppad \\
 & & $1.73 \times 10^{-9}$ & $1.64 \times 10^{-9}$ & $1.56 \times 10^{-9}$ & $1.55 \times 10^{-9}$ & $1.51 \times 10^{-9}$ & $1.50 \times 10^{-9}$ \toppad \\
\hhline{~|-|-|-|-|-|-|-|}
 & \multirow{2}{*}{10} & \gc{$5.16 \times 10^{-7}$} & \gc{$1.83 \times 10^{-7}$} & \gc{$2.64 \times 10^{-7}$} & \gc{$4.40 \times 10^{-7}$} & \gc{$5.15 \times 10^{-7}$} & \gcr{$2.66 \times 10^{-6}$} \toppad \\
 & & $1.90 \times 10^{-9}$ & $1.66 \times 10^{-9}$ & $1.60 \times 10^{-9}$ & $1.56 \times 10^{-9}$ & $1.54 \times 10^{-9}$ & $1.53 \times 10^{-9}$ \toppad \\
\hline
\end{tabular}
\end{footnotesize}
\end{table}

\begin{table}[htbp]
\caption{Comparison of analytic process models for the inverted pendulum system in \emph{Experiment~C2}. The table shows the comparison of the RMSE medians over 30~runs of the SNGP (grey) and MGGP (white) algorithm depending on the Gaussian noise standard deviation coefficient $\lambda$ and the number of training samples $n_s$.
}\label{tab:e3} \centering
\begin{footnotesize}
\begin{tabular}{|c|c|ccc|}
\hline
\multirow{2}{*}{Variable} & \multirow{2}{*}{$\lambda$} & \multicolumn{3}{c|}{Number of training samples $n_s$} \toppad \\
\hhline{~|~|-|-|-|}
 & & 20 & 100 & 1000 \toppad \\
\hline
\multirow{10}{*}{$\alpha$} & \multirow{2}{*}{0} & \gc{$9.58 \times 10^{-2}$} & \gc{$1.79 \times 10^{-2}$} & \gcr{$6.11 \times 10^{-3}$} \toppad \\
 & & $8.13 \times 10^{-2}$ & $1.05 \times 10^{-2}$ & $1.36 \times 10^{-2}$ \toppad \\
\hhline{~|-|-|-|-|}
 & \multirow{2}{*}{0.01} & \gc{$ 3.95 \times 10^{-1}$} & \gc{$1.45 \times 10^{-1}$} & \gcr{$2.80 \times 10^{-2}$} \toppad \\
 & & $3.96 \times 10^{-1}$ & $1.37 \times 10^{-1}$ & $2.85 \times 10^{-2}$ \toppad \\
\hhline{~|-|-|-|-|}
 & \multirow{2}{*}{0.05} & \gc{$1.15 \times 10^{0}$} & \gc{$4.89 \times 10^{-1}$} & \gcr{$1.43 \times 10^{-1}$} \toppad \\
 & & $8.69 \times 10^{-1}$ & $5.01 \times 10^{-1}$ & $1.42 \times 10^{-1}$ \toppad \\
\hhline{~|-|-|-|-|}
 & \multirow{2}{*}{0.1} & \gc{$1.90 \times 10^{0}$} & \gc{$7.61 \times 10^{-1}$} & \gcr{$3.54 \times 10^{-1}$} \toppad \\
 & & $2.26 \times 10^{0}$ & $8.22 \times 10^{-1}$ & $3.59 \times 10^{-1}$ \toppad \\
\hline
\multirow{10}{*}{$\dot\alpha$} & \multirow{2}{*}{0} & \gc{$4.56 \times 10^{0}$} & \gc{$7.65 \times 10^{-1}$} & \gcr{$5.04 \times 10^{-1}$} \toppad \\
 & & $3.89 \times 10^{0}$ & $7.56 \times 10^{-1}$ & $5.38 \times 10^{-1}$ \toppad \\
\hhline{~|-|-|-|-|}
 & \multirow{2}{*}{0.01} & \gc{$4.22 \times 10^{0}$} & \gc{$2.28 \times 10^{0}$} & \gcr{$8.13 \times 10^{-1}$} \toppad \\
 & & $4.71 \times 10^{0}$ & $2.75 \times 10^{0}$ & $8.18 \times 10^{-1}$ \toppad \\
\hhline{~|-|-|-|-|}
 & \multirow{2}{*}{0.05} & \gc{$7.89 \times 10^{0}$} & \gc{$6.07 \times 10^{0}$} & \gcr{$3.14 \times 10^{0}$} \toppad \\
 & & $7.39 \times 10^{0}$ & $6.75 \times 10^{0}$ & $2.76 \times 10^{0}$ \toppad \\
\hhline{~|-|-|-|-|}
 & \multirow{2}{*}{0.1} & \gc{$1.26 \times 10^{1}$} & \gc{$9.61 \times 10^{0}$} & \gcr{$6.65 \times 10^{0}$} \toppad \\
 & & $1.26 \times 10^{1}$ & $8.99 \times 10^{0}$ & $6.53 \times 10^{0}$ \toppad \\
\hline
\end{tabular}
\end{footnotesize}
\end{table}

\clearpage
\bibliographystyle{elsarticle-num}
\bibliography{asoc106432}

\end{document}